\documentclass[10pt,twocolumn,letterpaper]{article}

\usepackage{cvpr}              

\usepackage[pagebackref,breaklinks,colorlinks,citecolor=cvprblue]{hyperref}
\usepackage{times}
\usepackage{epsfig}
\usepackage{graphicx}
\usepackage{amsmath}
\usepackage{amssymb}
\usepackage{booktabs}
\usepackage{appendix}
\usepackage{multirow}
\usepackage{bbding}
\usepackage{makecell}
\usepackage{bm}
\usepackage[table]{xcolor}
\definecolor{cvprblue}{rgb}{0.21,0.49,0.74}
\usepackage[accsupp]{axessibility}
\def\paperID{126} 
\def\confName{CVPR}
\def\confYear{2024}

\title{HalluciDoctor: Mitigating Hallucinatory Toxicity in Visual Instruction Data}

\author{Qifan Yu$^1$ \quad Juncheng Li$^1$\footnotemark[2]\quad  Longhui Wei$^2$ \quad Liang Pang$^3$ \quad Wentao Ye$^1$ \quad Bosheng Qin$^1$ \\
Siliang Tang$^1$\quad\quad Qi Tian$^2$\quad\quad Yueting Zhuang$^1$\\ \small $^1$Zhejiang University, $^2$Huawei Cloud, $^3$	Institute of Computing Technology, Chinese Academy of Sciences\\ {\tt\small \{yuqifan, junchengli, bsqin, siliang, yzhuang\}@zju.edu.cn}\\{\tt\small \{weilonghui1, tian.qi1\}@huawei.com, pangliang@ict.ac.cn}}

\begin{document}
\maketitle
\renewcommand{\thefootnote}{\fnsymbol{footnote}} 
\footnotetext[2]{Juncheng Li is the corresponding author.}
\renewcommand{\thefootnote}{\arabic{footnote}}
\begin{abstract}
Multi-modal Large Language Models~(MLLMs) tuned on machine-generated instruction-following data have demonstrated remarkable performance in various multi-modal understanding and generation tasks. However, the hallucinations inherent in machine-generated data, which could lead to hallucinatory outputs in MLLMs, remain under-explored. This work aims to investigate various hallucinations~(\textsl{i.e.}, object, relation, attribute hallucinations) and mitigate those hallucinatory toxicities in large-scale machine-generated visual instruction datasets. Drawing on the human ability to identify factual errors, we present a novel hallucination detection and elimination framework, \textbf{HalluciDoctor}, based on the cross-checking paradigm. We use our framework to identify and eliminate hallucinations in the training data automatically. Interestingly, \textbf{HalluciDoctor} also indicates that spurious correlations arising from long-tail object co-occurrences contribute to hallucinations. Based on that, we execute counterfactual visual instruction expansion to balance data distribution, thereby enhancing MLLMs' resistance to hallucinations. Comprehensive experiments on hallucination evaluation benchmarks show that our method successfully mitigates 44.6\% hallucinations relatively and maintains competitive performance compared to LLaVA. The
data and code for this paper are publicly available.\footnote{\url{https://github.com/Yuqifan1117/HalluciDoctor}} 
 
\end{abstract}    
\section{Introduction}
\label{sec:intro}

Recently, Multi-modal Large Language Models~(MLLMs) have made promising progress on multi-modal tasks, such as image captioning, visual question-answering, and visual conversations~\cite{tsimpoukelli2021multimodal, li2022blip, li2023blip, alayrac2022flamingo}. Additionally, inspired by the impressive instruction-following capability of LLMs~\cite{touvron2023llama, chiang2023vicuna, peng2023instruction}, several more powerful MLLMs~\cite{liu2023visual, zhu2023minigpt, ye2023mplug, dai2023instructblip, li2023fine} have recently emerged, extending instruction-tuning to the multi-modal space. Due to the scarcity of visual-language instruction-following data, recent research\cite{liu2023visual, zhu2023minigpt} presents a data reformation approach, which leverages text-only LLMs conditioned on image captions and bounding boxes to create instruction-following data involving visual content.
\begin{figure}[!t]
    \centering
    \includegraphics[width=1.\linewidth]{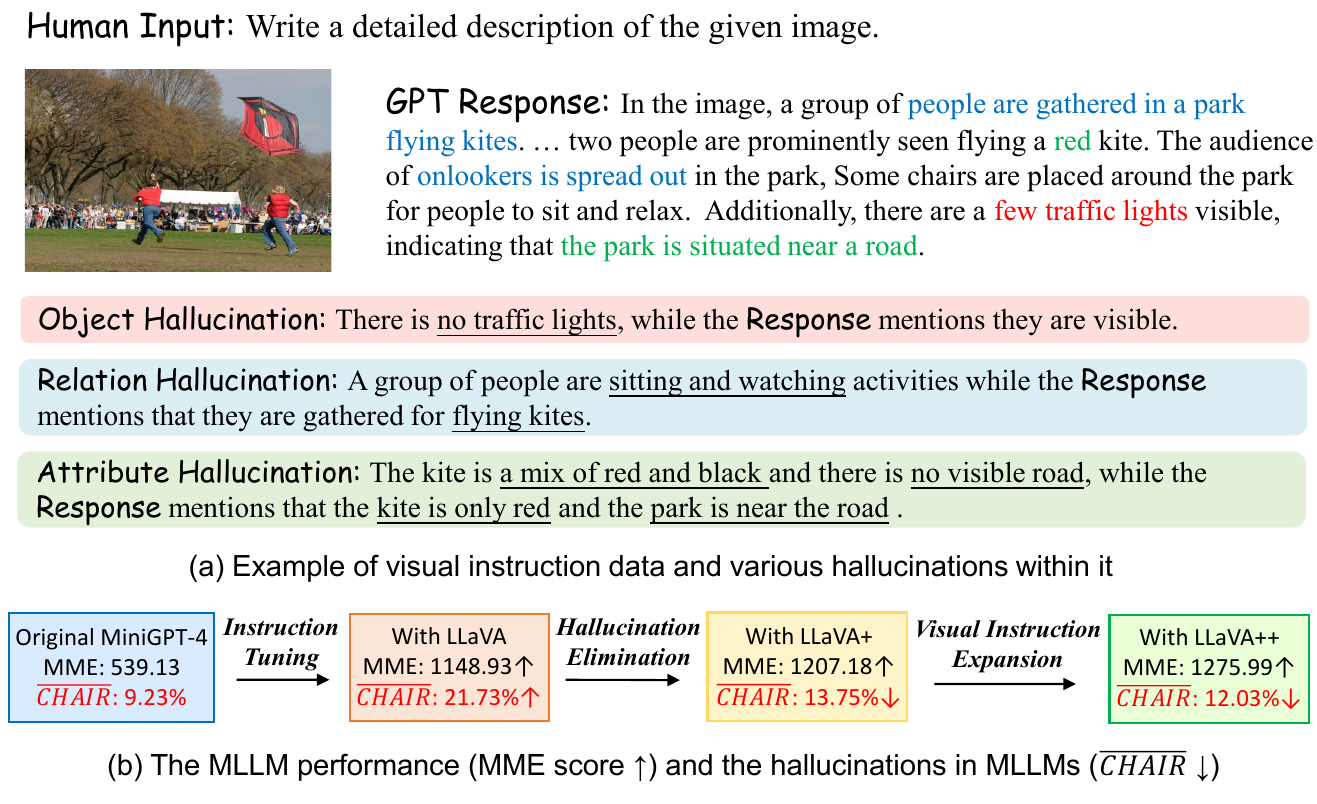}
    \caption{(a) On the top, we show an example of visual instruction and various hallucinatory toxicities within it. (b) At the bottom, we show that refined LLaVA++ from HalluciDoctor can alleviate hallucinatory toxicity to MLLM and improve its performance.} 
    \vspace{-4mm}
    \label{hallucination-example}
\end{figure}
However, these visual instructions might include hallucinatory responses incongruent with the image content, as they are produced by text-only large language models. For instance, as shown in Figure~\ref{hallucination-example}~(a), the response in the visual instruction data includes object hallucinations such as ``a few traffic lights" and relation hallucinations like ``people gathered for flying kites" instead of ``sitting and watching". These hallucinatory responses may compromise the MLLM's ability to perceive the real world accurately. 


Motivated by this insight, we systematically definite various kinds of hallucinations~(\textsl{i.e.}, object, relation, attribute hallucinations) and investigate them in visual instruction datasets. Using the widely used dataset LLaVA-Instruction-158K~\cite{liu2023visual}, we construct extended CHAIR metrics to comprehensively evaluate the impact of these visual instructions on modern MLLMs, considering both performance and hallucination issues. While instruction-tuning on LLaVA data improves MLLM performance, it significantly increases the probability of producing hallucinations~(Figure~\ref{hallucination-example}~(b), original MiniGPT-4 v.s. with LLaVA finetuning). These observations confirm with recent studies~\cite{liu2023aligning, gunjal2023detecting, li2023evaluating} that current machine-generated instruction-following data contains massive pernicious hallucinations~(32.6\% in LLaVA) that cause MLLM to produce inaccurate outputs. This pushes us to focus on mitigating those hallucinatory toxicities.

Previous works mainly focus on collecting extra remedial training data~\cite{liu2023aligning, wang2023vigc, lu2023evaluation} or utilize additional plug-in models~\cite{gunjal2023detecting, yin2023woodpecker, liao2023revo} to mitigate the hallucinations during inference. LRV-Instruction~\cite{liu2023aligning} proposes to add extra negative instructions to increase the robustness of the MLLM against hallucinations. M-HalDetect~\cite{gunjal2023detecting} mitigates hallucinations by incorporating an additional trained reward model during the inference phase. However, these methods either raise training labor costs or prolong inference time. Moreover, they merely superficially suppress the hallucinatory output of MLLMs, largely neglecting the inherent hallucinatory toxicity in the visual instruction dataset, which causes the hallucinatory errors in existing MLLMs. This leads to sub-optimal hallucination elimination for MLLMs.

In contrast to the methods above, we aim to eradicate hallucinations in machine-generated visual instruction data. The primary challenge is how to accurately detect and remove various hallucinations from massive such data without manual annotations. For this, we propose a flexible Hallucination Detection and Elimination Framework, namely \textbf{HalluciDoctor}, which automatically detects various hallucinations in arbitrary positions and dispels them based on a cross-checking paradigm. The key insight is that when asked about the hallucinatory content of a given image, the responses from different MLLM experts typically tend to vary and can even contradict each other. Specifically, as shown in Figure~\ref{framework}, HalluciDoctor breaks the hallucination detection procedure into three sub-processes: 1) \textit{Answer Chunks Extraction}: extract all answer chunks including objects, relations, and attributes by textual scene graph parsing as description-oriented answers; 2) \textit{Answer-based Question Generation}: generate corresponding fine-grained questions with diverse types for each answer; 3) \textit{Consistency Cross-Checking}: obtain image-oriented candidate answers from multiple MLLMs and cross-check the consistency between description-oriented answer chunks and their corresponding image-oriented answers. Subsequently, HalluciDoctor identifies those semantic chunks with consistency scores below a threshold as hallucinatory chunks. It eliminates these hallucination errors without disrupting the contextual semantics, resulting in the rectified dataset, LLaVA+. This significantly alleviates hallucinations in MLLMs~(Figure~\ref{hallucination-example}~(b), with LLaVA v.s. with LLaVA+).

In our exploration of eliminating hallucinations in visual instruction data, we find that HalluciDoctor not only assists in locating hallucinations but also indicates the spurious correlations causing them, which stem from the long-tail distribution of object co-occurrences. These spurious correlations can mislead MLLMs into erroneously inferring the presence of objects that do not exist in the images. Inspired by concepts of counterfactual generation~\cite{le2023coco, wu2023reasoning, zhang2023if, li2023large}, we propose a seesaw-based strategy for counterfactual visual instruction expansion. It resolves this issue by balancing the long-tail object co-occurrence distribution through two collaborative factors, ultimately creating a more robust visual instruction dataset, LLaVA++. 
This enables MLLMs to concentrate on accurately perceiving the content of images instead of spurious associations, thereby strengthening their resistance to hallucinations and overall performance~(Figure~\ref{hallucination-example}~(b), with LLaVA+ v.s. with LLaVA++). 

Our main contributions are summarized as follows:
\begin{itemize}
    \item To the best of our knowledge, we are the first to comprehensively investigate the severe hallucination toxicity in existing machine-generated visual instruction datasets.
    \item We propose a novel \textbf{HalluciDoctor} method to detect various hallucinations by a consistency cross-checking paradigm and dispel them in a low-resource way.
    \item Based on HalluciDoctor, we further automatically generate more counterfactual instruction data to improve the MLLMs' resistance to hallucinations.
    \item Our empirical study confirms our method's effectiveness in eliminating hallucinations in visual instruction data and improving MLLMs' robustness.

\end{itemize}

\section{Related works}
\label{relatedworks}

\subsection{Multi-modal Large Language Model} 
With the remarkable generalizability of LLMs in a zero-shot setting~\cite{zhao2023survey, ouyang2022training, wei2022chain, wang2022self}, early works integrating LLMs with visual modality have demonstrated impressive visual-language understanding ability~\cite{tsimpoukelli2021multimodal, li2023blip, yu2023visually, guo2023images, li2023otter}. Recently, more powerful MLLMs~\cite{yin2023survey, zhu2023minigpt, ye2023mplug, dai2023instructblip, li2023fine, bai2023qwen, gpt4} have emerged to mimic human perceptual capabilities for unseen vision-language tasks. Generally, MLLMs align the vision encoder into the LLM by a cross-modal alignment network~(\textit{e.g.}, a linear projection layer in MiniGPT-4~\cite{zhu2023minigpt}, a visual abstractor in mPLUG-Owl~\cite{ye2023mplug}, and Q-former in InstructBLIP~\cite{dai2023instructblip}). The training process of MLLMs mainly contains two stages: the first pre-training and the second multi-modal instruction tuning. Moreover, LLaVA~\cite{liu2023visual} leverages powerful LLMs to obtain extensive visual instruction data, paving the way for acquiring the instruction-following ability of MLLMs. This is essential for constructing more powerful MLLMs in a low-resource way.

\subsection{MLLM Hallucination}
Although MLLMs have demonstrated remarkable performances in various VL tasks, they still suffer from the hallucination phenomenon that textual outputs conflict with the visual content.
Current research on MLLM hallucinations mainly focuses on the detection and elimination of hallucinations~\cite{li2023evaluating, gunjal2023detecting, zhou2023analyzing, liu2023aligning, wang2023vigc, liao2023revo}. \cite{li2023evaluating} solely concentrates on object hallucinations and treats hallucination detection as a binary classification issue, limiting its evaluation for open-ended responses. HalDetect~\cite{gunjal2023detecting} identifies hallucinations by training a specialized classifier. But both methods need manual ground-truth answer collection and only consider the simplest object hallucinations.
For hallucination mitigation within MLLMs, previous works generally either collect more high-quality data manually~\cite{liu2023aligning} or attach an extra correction model~\cite{wang2023vigc, liao2023revo, gunjal2023detecting}. Furthermore, existing works solely concentrate on direct hallucinations in MLLM reasoning, ignoring the essential hallucinatory toxicity in the visual instruction data itself. Contrary to the methods above, we shift our attention to diverse hallucinations in the visual instruction data and devise an automated framework to detect and eliminate potential hallucinatory toxicity.


\section{The Toxicity of Visual Instruction Data}
Due to the scale limitation of multi-modal instruction-following data, there is a growing interest in self-generated instructions for MLLMs~\cite{liu2023visual, zhu2023minigpt}. However, since these visual instructions are generated by text-only GPT-4, they may contain numerous hallucinations, leading MLLMs to produce responses that inaccurately represent the images. To our knowledge, it is the first work to systematically analyze the hallucinatory toxicity of visual instruction datasets.
\subsection{Hallucination Metrics}
\label{metric}
To better analyze hallucinatory toxicity in the dataset, we first categorize three types of hallucinations frequently appearing in it: 1) \textit{object hallucination} is the object that appears in the description but not in the image. 2) \textit{relation hallucination} involves the relation between corresponding objects that exhibits inconsistency between descriptions and images. 3) \textit{attribute hallucination} refers to inaccurate object properties in the description, such as size, color, and states.

The current popular metric CHAIR~\cite{rohrbach2018object} only calculates the proportion of nonexistent objects in the description. Thus, we extend the naive CHAIR metric into more complex scenarios to evaluate various hallucinations. Initially, we incorporate synonym lists into annotated objects and phrases, forming an enhanced ground truth set. Subsequently, we split the description into sentences and extracted all objects, relations, and attributes for a comprehensive assessment. The extended $\text{CHAIR}$ metric then calculates the ratio of sentences containing hallucinatory elements not present in the image. Accordingly, the definition of extended CHAIR, including $\text{CHAIR}_{obj}$, $\text{CHAIR}_{rel}$, and $\text{CHAIR}_{attri}$, is delineated as follows:
\begin{eqnarray}
    \text{CHAIR}_{obj} = \frac{|\text{\{sentences\ with\ nonexistent\ object\}}|}{|\text{\{all \ sentences\}}|}
\end{eqnarray}
\begin{eqnarray}
    \text{CHAIR}_{rel} = \frac{|\text{\{sentences\ with\ nonexistent\ relation\}}|}{|\text{\{all\ sentences\}}|}
\end{eqnarray}
\begin{eqnarray}
    \text{CHAIR}_{attri} = \frac{|\text{\{sentences\ with\ nonexistent\ attribute\}}|}{|\text{\{all\ sentences\}}|}
\end{eqnarray}
The higher CHAIR score indicates there exist more hallucinations in the description. Notably, we compute the $\text{CHAIR}_{rel}$ and the $\text{CHAIR}_{attri}$ only among existent objects to avoid misjudging compositional errors influenced by object hallucinations as additional hallucination errors.
\subsection{Hallucinatory Toxicity Statistics}
Utilizing our extended CHAIR metric designed for in-depth hallucination analysis, we meticulously examine the hallucination frequency within machine-generated visual instruction data. We concentrate on prevalent, machine-generated visual instruction datasets, namely LLaVA~\cite{liu2023visual} and MiniGPT4-Instruction~\cite{zhu2023minigpt}.
LLaVA consists of 158K distinct instruction-following samples generated by GPT-4~\cite{gpt4}, while MiniGPT4-Instruction includes about 3.5K instances refined by ChatGPT~\cite{chatgpt} from detailed descriptions. To tackle the challenge of incomplete annotations in the above datasets, particularly regarding relations and attributes, we adopt GroundingDINO~\cite{liu2023grounding} to annotate objects and use the image-text similarity of BLIP~\cite{li2022blip} to judge the existence of relations and attributes. By incorporating these pseudo-labels, we thoroughly assess the hallucinatory toxicity in those datasets, as depicted in Table~\ref{hallucination_statistics}.
It can be seen that machine-generated visual instruction data uniformly exhibit various distinct types of hallucinations. However, the proxy detection approach lacks flexibility in addressing a variety of unidentified hallucinations due to its reliance on accurate annotation.
Therefore, we propose a general hallucination detection and elimination framework \textbf{HalluciDoctor} based on the consistency cross-checking paradigm to handle the potential hallucinations in the training data.

\begin{table}[t]
\resizebox{0.47\textwidth}{!}{
\centering
\begin{tabular}{lccccc}
\hline
\toprule
Dataset & \#Samples & $\text{CHAIR}_{obj}$~$\downarrow$ & $\text{CHAIR}_{rel}$~$\downarrow$ & $\text{CHAIR}_{attri}$~$\downarrow$ & Length\\ \hline
LLaVA~\cite{liu2023visual} & 158K & 28.1 & 36.0 & 33.7 & 96.1 \\ 
\textbf{LLaVA+} & 158K & \textbf{8.3} & \textbf{20.7} & \textbf{17.1} & 87.8 \\\hline
MiniGPT4-Instruction~\cite{zhu2023minigpt}& 3.5K & 22.6 & 35.6 & 31.6 & 70.8 \\
\textbf{MiniGPT4-Instruction+} & 3.5K & \textbf{13.3} & \textbf{21.7} & \textbf{23.8} & 61.8 \\ 

\bottomrule
\end{tabular}
}
\caption{The statistics of three types of hallucinations in visual instruction datasets and comparison with their corresponding rectified version by HalluciDoctor~(\textbf{bolded rows}).}
\vspace{-4mm}
\label{hallucination_statistics}
\end{table} 

\begin{figure*}
  \centering
    \includegraphics[width=1.\linewidth]{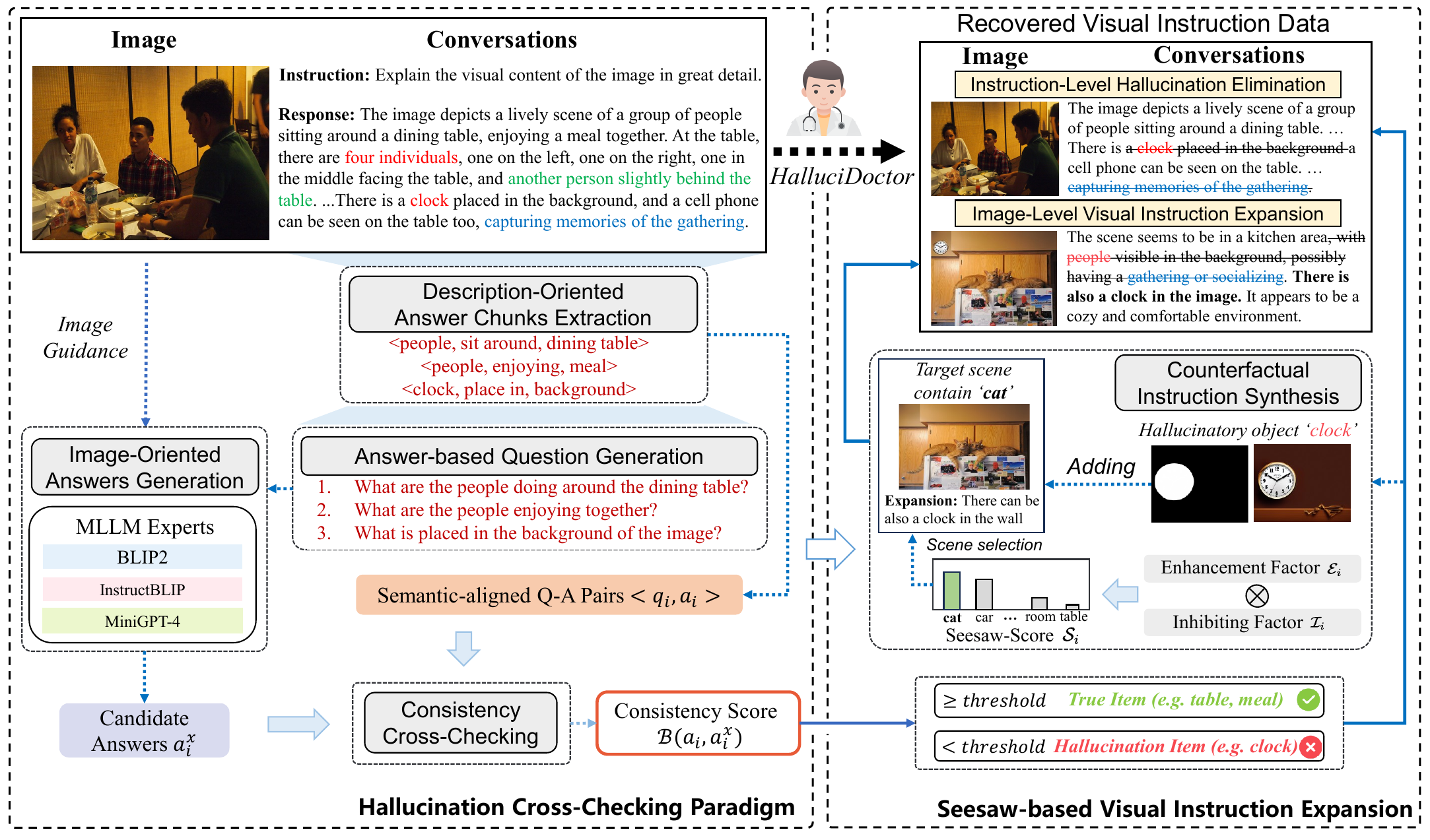}
  \caption{Overview of our proposed HalluciDoctor for automatically eliminating hallucinatory toxicity in visual instruction data and enhancing MLLM's resistance to hallucinations. We summarize the process into four steps: (1) HalluciDoctor first extracts description-oriented answers for semantic analysis and formulates corresponding questions. (2) Image-oriented candidate answers for these questions are then gathered from various MLLMs. (3) HalluciDoctor will identify and remove hallucinatory chunks via consistency cross-checking. (4) Lastly, It creates counterfactual instructions guided by preceding steps to expand the dataset and mitigate hallucinations radically.}
  \vspace{-5mm}
  \label{framework}
\end{figure*}

\section{HalluciDoctor Framework}
As illustrated in Figure~\ref{framework}, our HalluciDoctor framework consists of two primary modules. \textbf{Hallucination Cross-Check Paradigm} is designed to probe and eliminate hallucinatory errors in the original visual instruction data. \textbf{Seesaw-based Visual Instruction Expansion} produces additional counterfactual visual instructions to reduce the hallucinatory effects caused by spurious correlations and strengthen the MLLMs' resistance to hallucinations.
\subsection{Hallucination Cross-Checking Paradigm}
\label{CCF_1}

\noindent\textbf{Overview.} To exhaustively identify various types of hallucinations~(\textsl{i.e.}, object hallucinations, relation hallucinations, attribute hallucinations) in the corresponding description of each image, we introduce a hallucination cross-checking paradigm. Our insight is to decompose this demanding detection task into several simpler answer consistency-checking tasks. This paradigm comprises three subtasks: answer chunks generation, question generation, and consistency cross-checking. Then, HalluciDoctor eliminates any detected hallucinations. More details are in Appendix \textcolor{red}{B}.



\noindent\textbf{Answer Chunks Generation.} Since the generated description contains massive concepts, we employ a textual scene graph parser~\cite{li2023factual} to extract description-oriented answer chunks to represent the specific semantics, including objects, attributes, and relations. Given an image $I$ with its instruction-following data~$(\rm{X}_q, \rm{X}_a)$, where $\rm{X}_q$ is the instruction from human and $\rm{X}_a$ is the generated description from GPT-4. 
We extra all answer chunks from instruction-following data $(\rm{X}_q, \rm{X}_a)$ together as follows,
\begin{equation}
   A =\{a_{1},...,a_{n}\}
\end{equation}
where $a_i$ represents the $i^{th}$ answer chunk in the $(\rm{X}_q, \rm{X}_a)$.

\noindent\textbf{Question Generation.} Upon generating answer chunks, we construct corresponding questions, which are then used to derive image-oriented candidate answers. We employ LLM, like ChatGPT as potent question generators to encompass the wide diversity of answer chunks and question types. Thus, we can generate questions tailored to different answers, including concrete objects, abstract relations, and attribute descriptions as follows,
\begin{equation}
   Q =\{q_{1},...,q_{n}\}
\end{equation}
where $q_i$ represents the $i^{th}$ question corresponding to $a_i$.

\noindent\textbf{Consistency Cross-Checking.} The consistency cross-checking step aims to verify the consistency between each answer chunk $a_i$ and its corresponding content in the image. 
To obtain the specific image content, we use the generated questions to derive image-oriented candidate answers based on MLLM experts~(\textit{e.g.}, BLIP2~\cite{li2023blip}, InstructBLIP~\cite{dai2023instructblip}, and MiniGPT-4~\cite{zhu2023minigpt}). Formally, let $\{\mathcal{F}_x\}$ denote an MLLM expert, the prompt template slotted with the reference image $I$ and one question $q_i$ is fed to the MLLM $\mathcal{F}_x$ to produce the image-oriented candidate answer $a_i^x=\mathcal{F}_x(I,q_i)$. Once we collect image-oriented candidate answers $\{a_i^x\}$ for each answer chunk $a_i$, we will compare their consistency. We employ a Bert-based metric~(\textsl{i.e.}, BEM~\cite{bulian2022tomayto}) to evaluate their consistency since this metric provides more flexibility in the answer formulation than strictly hard matching. 
Let $\mathcal{B}(\cdot,\cdot|q)$ denote the BEM score between two answers according to the question, the final \textit{ConScore} of the answer chunk $a_i$ is calculated by voting as follows,
\begin{equation}
    ConScore_i = \frac{1}{m}\sum_{x=1}^m\mathcal{B}(a_i,a_i^x)
\end{equation}
where $m$ denotes the number of MLLM experts. We will consider the answer chunk $a_i$ as a hallucination when its ConScore $<$ 0.5. More discussion of BEM evaluation and threshold determination is presented in Appendix \textcolor{red}{B.2}.


\noindent\textbf{Hallucination Elimination.} Benefiting from the above subtasks, we can accurately locate hallucinatory chunks with their contexts. We employ ChatGPT to automatically remove the hallucinatory chunks based on the context and guarantee the coherence and harmony of the corresponding sentences. After this rectification process, we obtain more accurate visual instruction data by diminishing erroneous hallucinatory descriptions, denoted as LLaVA+. This step is designed to alleviate the hallucinatory toxicity in machine-generated visual instruction data for MLLM training.

\begin{table*}[!t]
\resizebox{0.99\textwidth}{!}{
\begin{tabular}{lclccc|ccc}
\hline
\toprule
\multicolumn{2}{c}{\multirow{2}{*}{Model Type}} &\multicolumn{1}{c}{\multirow{2}{*}{Methods}} & \multicolumn{3}{c}{Instance-level} &\multicolumn{3}{c}{Sentence-level} \\\cline{4-6} \cline{7-9}
& &  & $\text{CHAIR}_{obj}$ $\downarrow$ & $\text{CHAIR}_{rel}$ $\downarrow$ &$\text{CHAIR}_{attri}$ $\downarrow$ & $\text{CHAIR}_{obj}$ $\downarrow$ & $\text{CHAIR}_{rel}$ $\downarrow$ & $\text{CHAIR}_{attri}$ $\downarrow$ \\\hline
\multicolumn{2}{c}{\multirow{3}{*}{Specific}}& Faithful Prompt & 9.3& 11.1 & 14.1& 23.2 & 24.8 & 25.5 \\
& &LURC\cite{zhou2023analyzing} & 5.7& 7.6 & 13.3& 16.0 & 22.8 & 28.5 \\
& &VIGC\cite{wang2023vigc} & 6.1& 7.5 & 11.5& 15.2 & 17.7 & 22.3 \\ \hline
\multirow{8}{*}{\rotatebox[]{90}{Model-Agnostic}}&\multirow{4}{*}{MiniGPT4~(7B)} & w/ LLaVA~\cite{liu2023visual} & 12.0 & 12.2 & 10.1 & 35.0 &34.8&26.3\\ 
& & w/ LRV~\cite{liu2023aligning} &10.0 & 10.8&13.6&24.9& 21.0 &24.8\\ 
& & \cellcolor{gray!30}w/ LLaVA+ & \cellcolor{gray!30}5.9 & \cellcolor{gray!30}6.1 & \cellcolor{gray!30}8.5 & \cellcolor{gray!30}19.6 &\cellcolor{gray!30}20.5& \cellcolor{gray!30}21.9\\
& & \cellcolor{gray!30}w/ LLaVA++ &\cellcolor{gray!30}\textbf{5.7} &\cellcolor{gray!30}\textbf{5.4}&\cellcolor{gray!30}\textbf{7.1}&\cellcolor{gray!30}\textbf{16.6} &\cellcolor{gray!30}\textbf{17.1} &\cellcolor{gray!30}\textbf{20.3} \\\cline{2-9}
& \multirow{4}{*}{mPLUG-Owl~(7B)}& w/ LLaVA~\cite{liu2023visual} & 10.6& 10.0&10.3&32.6&32.0& 29.1 \\ 
& & w/ LRV~\cite{liu2023aligning} & 10.3&9.5 &13.0&30.8& 29.6&32.1\\ 
& & \cellcolor{gray!30}w/ LLaVA+ & \cellcolor{gray!30}7.6 & \cellcolor{gray!30}7.1 & \cellcolor{gray!30}8.0 & \cellcolor{gray!30}22.2& \cellcolor{gray!30}21.1 & \cellcolor{gray!30}24.2 \\ 
& & \cellcolor{gray!30}w/ LLaVA++ & \cellcolor{gray!30}\textbf{6.4} & \cellcolor{gray!30}\textbf{5.5}&\cellcolor{gray!30}\textbf{6.7}&\cellcolor{gray!30}\textbf{19.3}& \cellcolor{gray!30}\textbf{17.6}&\cellcolor{gray!30}\textbf{16.5}\\ 
\bottomrule
\end{tabular}
}
\caption{Comprehensive CHAIR evaluation results to show the recovery effect of hallucination elimination by HalluciDoctor for MLLMs.}
\label{hallucination_chair}
\vspace{-0.4cm}
\end{table*} 
\subsection{Seesaw-based Visual Instruction Expansion}
\label{CCF_3}
In our efforts to locate hallucinations within visual instruction data, we have observed a notable pattern: these hallucinations frequently occur alongside objects that often appear together due to the long-tail distribution of object co-occurrences. For instance, in visual datasets, commonly co-occurring objects like `cars' and `roads' may inadvertently lead to spurious correlations, resulting in hallucinations. Consequently, a model trained on such data might incorrectly infer the presence of a `car' when encountering a `road' even in images where such a pairing is absent. This misunderstanding is the root cause of many hallucinations in visual data. 
By integrating hallucinatory objects into tail scenes where they rarely appear, we introduce counterfactual interventions~\cite{pawlowski2020deep} to mitigate the spurious correlations among strongly associated objects.

However, selecting such scenes is challenging, as it requires balancing the rarity of co-occurrence with the hallucinatory objects and their contextual plausibility. To this end, we propose a seesaw-based strategy with an enhancement factor and an inhibiting factor to adaptively select target scenes for counterfactual instruction expansion.

\noindent\textbf{Enhancement Factor.} 
Given an illusory object $o$ in the response, we denote the object in the response that appears most frequently with $o$ across all annotations as $o^*$ and their co-occurrence frequency as $n^*$.
The enhancement factor $\mathcal{E}_{i}$ is designed to increase the weight of other objects $o_i$ that rarely co-occur with $o$ and is computed as follows,
\begin{equation}
    \mathcal{E}_{i}=\left\{
	\begin{array}{ccc}
		\frac{n^*}{\max(n_i,1)}      &  ,   & {\rm if}\ {n_i\leq n^*}\\
		1    &  ,   & {\rm if}\ {n_i>n^*}\\ 
	\end{array}\right.~.
\label{eq7}
\end{equation}
where $n_i$ represents the co-occurrence frequency between the hallucinatory object $o$ and other object $o_i$, and $\mathcal{E}_{i}$ is inversely correlated with their co-occurrence frequency.

\noindent\textbf{Inhibiting Factor.} 
Although the enhancement factor effectively promotes the co-occurrence of the hallucinatory object with its infrequently co-occurring objects, it overlooks the contextual plausibility of these co-occurring pairs. Therefore, we introduce the inhibiting factor $\mathcal{I}_{i}$ to suppress the weight of objects in rare combinations with the contextually relevant object $o^*$ as follows,
\begin{equation} 
    \mathcal{I}_{i}=\left\{
	\begin{array}{ccc}
		\frac{m_i}{n^*}    &  ,   & {\rm if}\ {m_i\leq n^*}\\ 
                1      &  ,   & {\rm if}\ {m_i>n^*}\\
	\end{array}\right.~.
\end{equation}
where $m_i$ represents the co-occurrence frequency between the $o$'s most contextually relevant object $o^*$ and other objects $o_i$. Once we obtain $\mathcal{E}_{i}$ and $\mathcal{I}_{i}$, we calculate the Seesaw-Score $\mathcal{S}_{i}$ as demonstrated below.
\begin{equation} 
    \mathcal{S}_{i}=\mathcal{E}_{i}*\mathcal{I}_{i}~.
\label{eq9}
\end{equation}
In eq.~\ref{eq7}-\ref{eq9}, rare objects exhibit a higher $\mathcal{E}$, while reasonable combinations show a higher $\mathcal{I}$. In this way, we identify scenes containing objects with the highest seesaw scores as target scenes and integrate $o$ into these scenes. We phrase its description as \texttt{There is also a/an \{object\} in the image} and create counterfactual instructions. 
Subsequently, we incorporate hallucinatory objects into suitable locations of target scenes guided by the bounding box, to facilitate the corresponding counterfactual image synthesis. The generated counterfactual instructions are then amalgamated with the rectified dataset, LLaVA+, to form a more robust dataset for MLLM instruction tuning, LLaVA++, thereby reducing the impact of spurious correlations on hallucinations. More details and analyses are in Appendix \textcolor{red}{B.4}.

\section{Experiments}
\label{session4}
In this section, we present both qualitative and quantitative experimental results and corresponding analyses to assess HalluciDoctor's superiority. Our focus primarily lies on detailed experiments regarding MLLM hallucinations~(\S~\ref{hallucinations}) and MLLM performance~(\S~\ref{performance}), and based on these aspects, we organize more comprehensive GPT-4 evaluations~(\S~\ref{gpt4_evaluation}) and human assessments~(\S~\ref{human_eval}) to evaluate MLLMs' open-ended capabilities precisely. 

\subsection{Experimental Setup}

\noindent\textbf{Model Setting.} In this paper, we utilize the most widely used machine-generated visual instruction data LLaVA-158K~\cite{liu2023visual} to conduct experiments. 
To provide a comprehensive evaluation, we thoroughly compare our HalluciDoctor~(\textit{w/ LLaVA+} and \textit{w/ LLaVA++}, shown in the \raisebox{0.7ex}{\colorbox{gray!40}{}} of Table~\ref{hallucination_chair}) with various SOTA methods tailored for alleviating hallucinations. We categorize those methods into two categories: (1) Specialized approaches. We incorporate some models requiring additional dedicated modules to mitigate hallucinations, including LURC\cite{zhou2023analyzing} and VIGC~\cite{wang2023vigc}, as well as employing explicit faithful prompts to constrain the generation of reliable instruction data~(Faithful Prompt). (2) Model-agnostic baselines. They refer to plug-and-play methods for optimization at the dataset level and fine-tune MLLMs on the corresponding instruction data, including \textit{w/ LLaVA}~\cite{liu2023visual} and \textit{w/ LRV}~\cite{liu2023aligning}. For model-agnostic baselines, we equip variant datasets with two popular MLLMs: MiniGPT-4~\cite{zhu2023minigpt} and mPLUG-Owl~\cite{ye2023mplug}. We use the official pre-trained MLLM with the image-text alignment stage and only fine-tune it in the second stage for fair comparison.


\subsection{Comparion of MLLM Hallucinations}
To assess the toxicity of visual instruction data on MLLMs and the efficacy of Hallucidoctor in eliminating hallucinations, we compared MLLMs with Hallucidoctor against baseline models using our extended CHAIR benchmark. In addition to the sentence-level CHAIR evaluation mentioned above~(\textsl{i.e.}, $\text{CHAIR}_S$), we further computed CHAIR at the instance-level~(\textsl{i.e.}, $\text{CHAIR}_I$) by quantifying all nonexistent instances within a sentence to assess the overall distribution of hallucinations. Following previous works~\cite{liu2023aligning, li2023evaluating}, we randomly select 500 unique images from the intersection of MSCOCO~\cite{lin2014microsoft} and Visual Genome~\cite{krishna2017visual} for a more detailed evaluation. Notably, these images were different from the ones used in LLaVA-158k and contained various kinds of annotations. Subsequently, we prompt MLLMs with the instruction of \texttt{Provide a detailed description of the given image} to generate detailed captions of similar length. We compute $\text{CHAIR}_{obj}$, $\text{CHAIR}_{rel}$, and $\text{CHAIR}_{attri}$ at two different levels and report the results in Table~\ref{hallucination_chair}. Based on the observation of experimental results, we have summarized the following conclusions:

\textbf{Visual instruction data has serious hallucinatory toxicity.} We compare the MLLM fine-tuned with LLaVA against the original MLLM. The former is more susceptible to generating hallucinations, particularly in object perception. This confirms our previous analysis of the hallucinatory toxicity in visual instruction data, further emphasizing the necessity to eliminate hallucinations therein.

\textbf{Our HalluciDoctor can be flexibly equipped to
different MLLMs for hallucination elimination.} We utilize HalluciDoctor to eliminate hallucinations and obtain the LLaVA+. For hallucination evaluation, we integrate this rectified dataset into two backbone models, MiniGPT-4~\cite{zhu2023minigpt} and mPLUG-Owl~\cite{ye2023mplug}. Despite the model diversity, MLLMs with LLaVA+ can consistently reduce the probability of various hallucinations~(average reduction of 4.6\% / 11.4\% in MiniGPT-4 and 2.7\% / 8.7\% in mPLUG-Owl in two metric levels). The results confirm that our HalluciDoctor, by effectively reducing hallucination errors in visual instruction data (as shown in the bolded rows in Table~\ref{hallucination_statistics}), can alleviate hallucinatory outputs in MLLMs, thereby enhancing their reliability in the real world.

\textbf{Compared with other model-agnostic methods, our
HalluciDoctor outperforms all of them for hallucination elimination.} Specifically, MLLMs fine-tuned on LLaVA+ exhibit fewer hallucinations among all three types compared to those trained on the meticulously curated SOTA dataset LRV-Instruction~\cite{liu2023aligning}, especially for more challenging attribute hallucinations~(8.5\% v.s. 13.6\% in instance-level of MiniGPT-4). Similar results are also observed in the closed-ended POPE evaluation~\cite{li2023evaluating}, with detailed analysis in Appendix \textcolor{red}{C.2}. It indicates that our HalluciDoctor effectively eliminates massive hallucinations in machine-generated data, constructing higher-quality visual instructions to mitigate hallucinatory toxicity in MLLMs.  


\textbf{Visual instruction expansion can effectively reduce hallucinations caused by spurious correlations.} With the help of LLaVA++, MLLM obtained fewer object hallucinations than LLaVA+~(17.1\% v.s. 20.5\% in sentence-level $\text{CHAIR}_{rel}$). It suggests that expanded counterfactual instructions can equalize the long-tail distribution of object co-occurrences, reducing MLLMs' inclination towards incorrect associations. Remarkably, the MLLM fine-tuned on LLaVA++ exhibited the fewest hallucinations across all metrics, highlighting the superiority of HalluciDoctor in enhancing model reliability. 
To further mitigate the impact of long-tail distributions, it is also promising to incorporate contrastive learning~\cite{zhang2022incorporating} with counterfactual interventions. 

\begin{table}[!t]
\resizebox{0.475\textwidth}{!}{
\begin{tabular}{lclcc}
\hline
\toprule
&Model Type & \multicolumn{1}{c}{Methods} &  Perception & Cognition \\ \hline

\multirow{11}{*}{\rotatebox[]{90}{MME Benchmark}} &\multirow{3}{*}{Specific}& Faithful Prompt & 696.34 & 293.57 \\
&&LURC\cite{zhou2023analyzing} & 904.61 & 254.13\\
&&VIGC\cite{wang2023vigc} & 879.35 & 221.79 \\ \cline{2-5}
&\multirow{4}{*}{MiniGPT4~(7B)} & w/ LLaVA~\cite{liu2023visual} & 859.64 & 289.29\\ 
& & w/ LRV~\cite{liu2023aligning} & 870.12 & 291.79\\ 
& & w/ LLaVA+ & 889.32 & 317.86\\
& & w/ LLaVA++ & \textbf{955.28} & \textbf{320.71} \\\cline{2-5}
&\multirow{4}{*}{mPLUG-Owl~(7B)}& w/ LLaVA & 967.34 & 276.07\\ 
& & w/ LRV~\cite{liu2023aligning} & 1008.57  & 263.93 \\
& & w/ LLaVA+ &1043.19 &282.14\\ 
& & w/ LLaVA++ & \textbf{1114.34} & \textbf{302.86} \\ 
\bottomrule
\end{tabular}
}
\caption{Results on MME evaluation~\cite{fu2023mme} of MiniGPT4-7B and mPLUG-Owl-7B. The performance is measured by the sum of the subtasks' scores, where the best score for each partition is bolded.}
\label{mme_evaluation}
\vspace{-0.4cm}
\end{table} 


\label{hallucinations}
\subsection{Comparion of MLLM Performance}
\label{performance}
While the extended CHAIR evaluation affirms HalluciDoctor's efficacy in hallucination elimination, a well-rounded analysis of its impact on MLLM performance remains to be conducted. Therefore, we conduct quantitative analysis on the MME benchmark~\cite{fu2023mme}, which evaluates the perception and cognition abilities of MLLMs on 14 subtasks. This setup converts human annotations into a series of "yes or no" questions and measures MLLM performance by calculating the total accuracy score. Table~\ref{mme_evaluation} summarizes the cognitive and perceptual performance of MLLMs fine-tuning on different datasets. 
Compared to LLaVA, LLaVA+ not only mitigates hallucinations but also achieves higher MLLM performance~(1207.18 v.s. 1148.93). In comparison to LRV-Instruction~\cite{liu2023aligning}, LLaVA+ performs better than this SOTA method, even with fewer visual instructions. This indicates that HalluciDoctor still preserves the accurate elements when eliminating hallucinations in visual instructions for better instruction alignment.
Additionally, LLaVA++ offers more challenging counterfactual instructions for better generalization. With the aid of LLaVA++, both fine-tuned MLLMs experience further improvement than LLaVA~(+127.06 / +173.79 in overall performance). 

\subsection{GPT-4 Evaluation}
\label{gpt4_evaluation}
\begin{figure}[t]
  \centering
    \includegraphics[width=1.\linewidth]{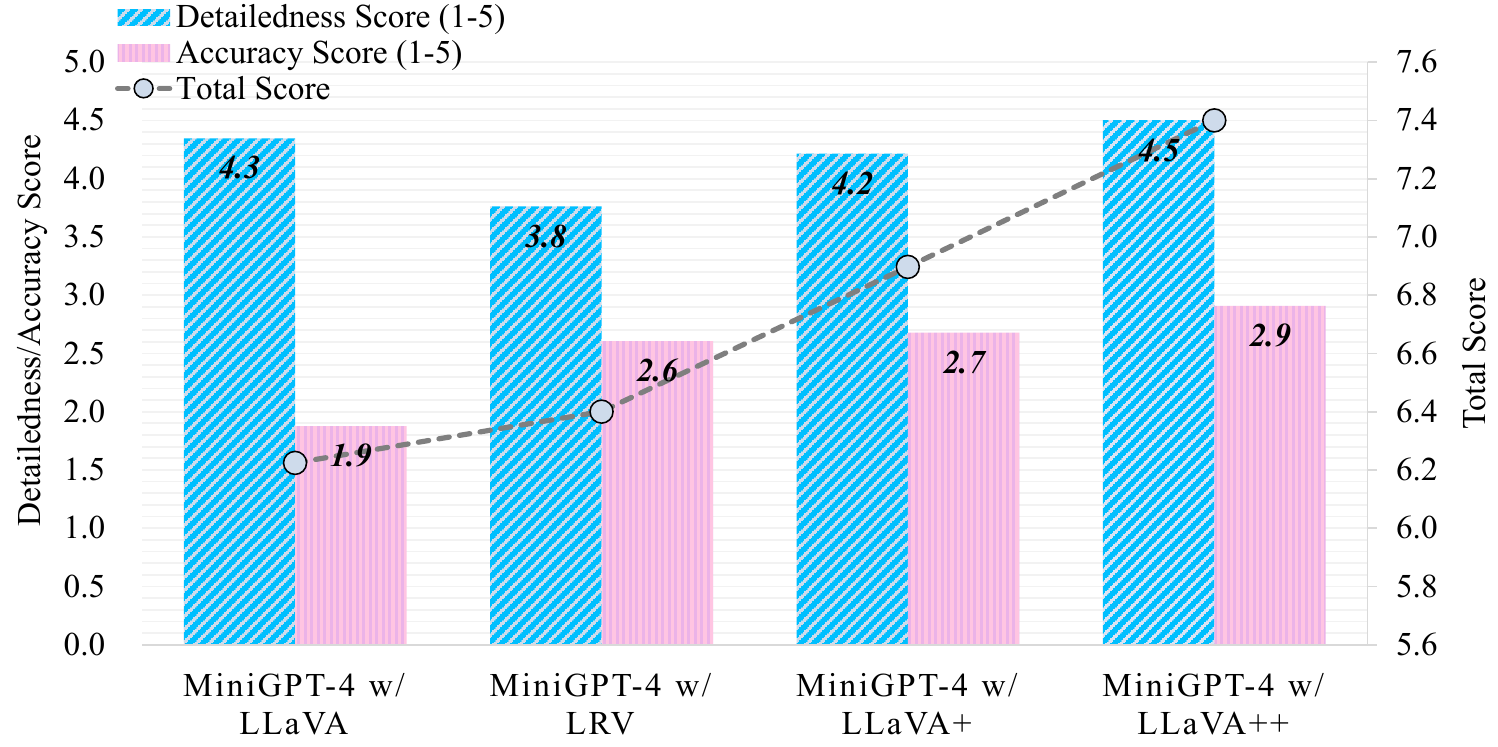}
    \vspace{-6mm}
  \caption{Evaluation scores of detailedness and accuracy for descriptions from MiniGPT-4 with different setups. We visualized the total scores using a gray line, where higher scores indicate more detailed descriptions and fewer hallucinations.}
  \vspace{-4mm}
  \label{gpt4_results}
\end{figure}
Empirically, there is a trade-off between the diversity of descriptions generated by MLLMs and the frequency of hallucinations. To strike a balanced evaluation between hallucinations and performance, we employ GPT-4 for a comprehensive assessment, comparing detailedness and accuracy in descriptions generated by MLLMs. In Figure~\ref{gpt4_results}, we visualize the GPT-4 evaluation scores under different MLLM settings. On the one hand, LLaVA significantly enriches descriptive content but at the cost of increased hallucinations. On the other hand, LRV-Instruction~\cite{liu2023aligning} reduces such hallucinatory errors at the expense of descriptive diversity, thus constraining the generative potential of the model. In contrast, our HalluciDoctor outperforms other models on a total score, indicating that it can eliminate hallucinations while preserving MLLM's ability to output diverse descriptions.
\begin{figure*}[!t]
  \centering
    \includegraphics[width=1.\linewidth]{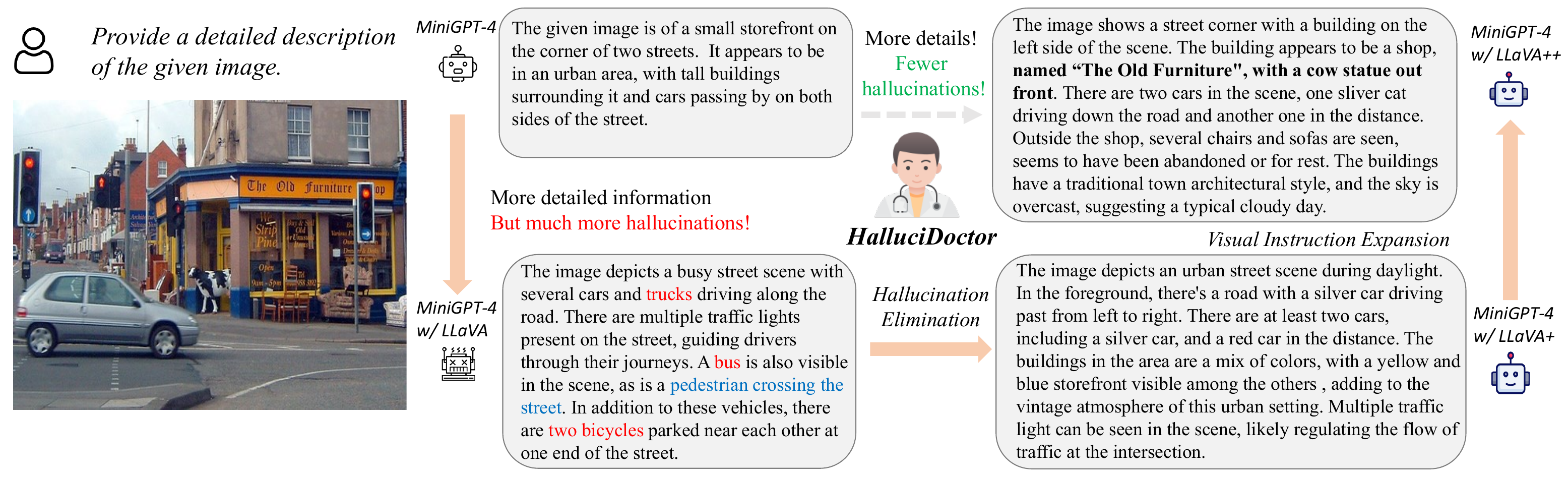}
    \vspace{-6mm}
  \caption{A case study comparing the levels of MLLM hallucination after fine-tuning on various instruction data.}
  \label{visualization}
  \vspace{-5mm}
\end{figure*}
\subsection{Human Evaluation}
\label{human_eval}

\begin{figure}[t]

  \centering
   \begin{subfigure}{0.495\linewidth}
    \centering
    \includegraphics[width=1.\linewidth]{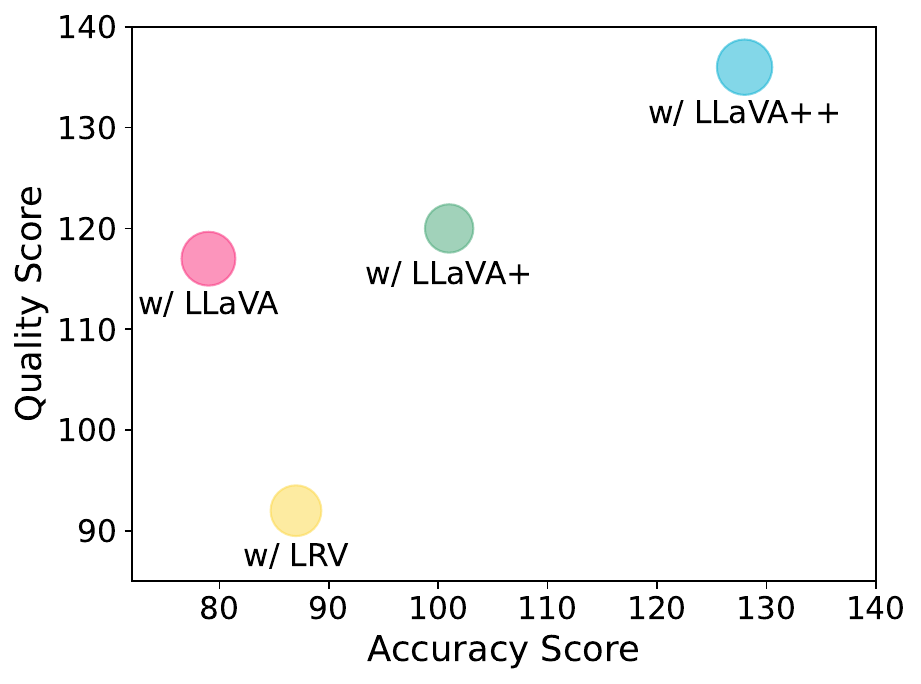}
    \caption{MiniGPT-4 Results}
    \label{input}
  \end{subfigure}
  \begin{subfigure}{0.495\linewidth}
    \centering
    \includegraphics[width=1.0\linewidth]{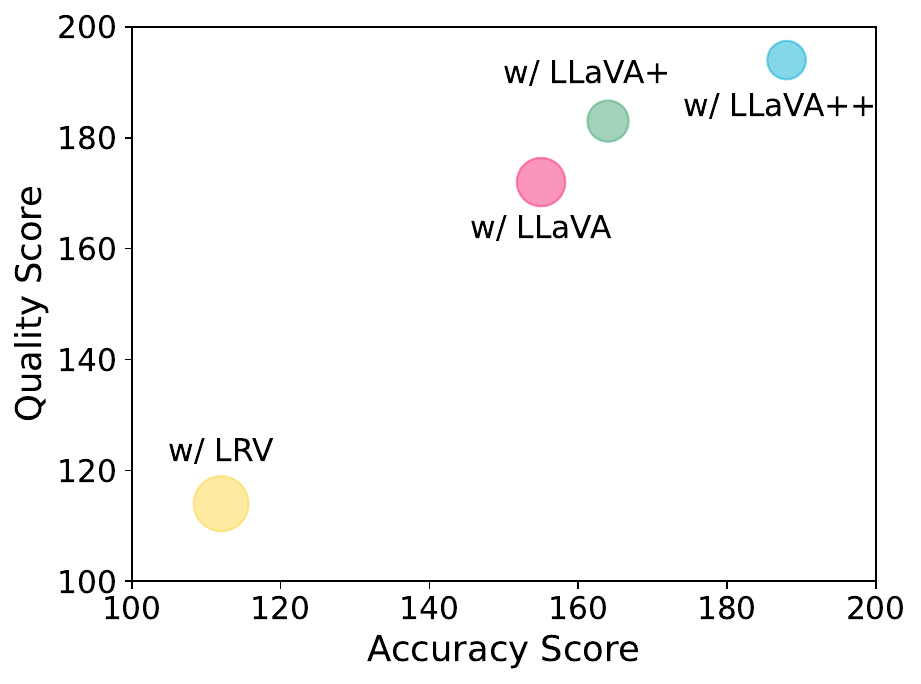}
    \caption{mPLUG-Owl Results}
    \label{sg1}
  \end{subfigure}
  \caption{Quality score~(y-axis, higher is better), accuracy score~(x-axis, higher is better), and the stability~(circle sizes, smaller is better) of MLLMs' responses on OwlEval benchmark.}
  \vspace{-5mm}
  \label{owleval}
\end{figure}

To more comprehensively assess the open-ended capabilities of MLLMs, we further conduct human evaluations using the OwlEval benchmark. \textbf{OwlEval}~\cite{ye2023mplug} is another open-ended evaluation set with 82 artificially constructed questions. We quantified responses from all models on the 3-0 scale (aligned with option A-D in the official setting), calculating quality and accuracy scores based on the relevance of the response to the question and the precision of the description, respectively. Additionally, we computed the score variance among each MLLM's responses to evaluate model stability. We show the visualized results for all MLLMs in Figure~\ref{owleval}. We observe that MLLMs fine-tuned with LLaVA++ achieve the highest accuracy scores regarding image content while maintaining fidelity to the corresponding questions. This suggests that HalluciDoctor effectively reduces hallucinations without compromising response quality, adeptly addressing a wide range of open-domain questions rather than limiting response length. 
\subsection{In-depth Analysis}



\begin{figure}
  \centering
   \begin{subfigure}{0.495\linewidth}
    \centering
    \includegraphics[width=1.\linewidth]{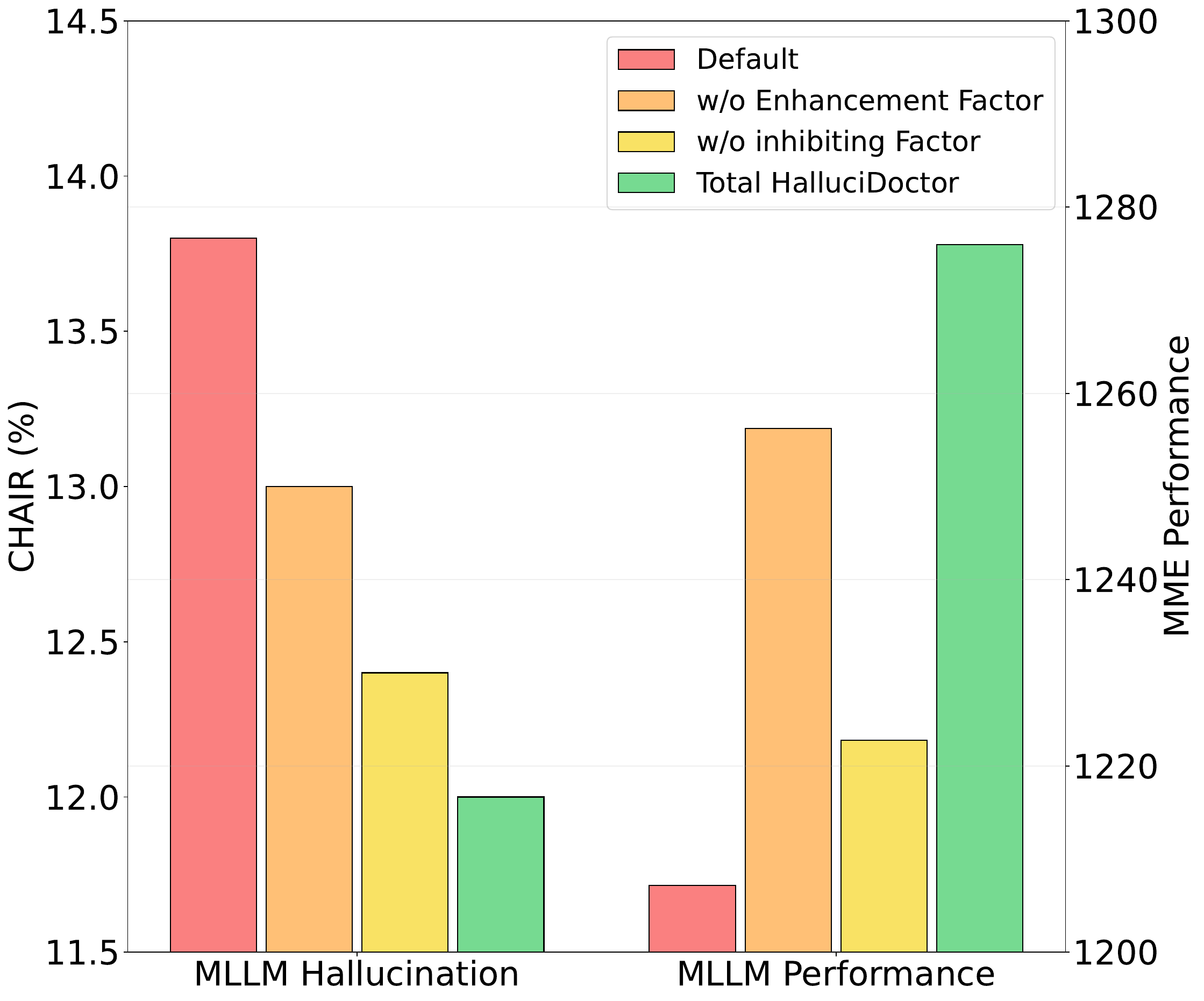}
    \caption{MiniGPT-4}
  \end{subfigure}
  \begin{subfigure}{0.495\linewidth}
    \centering
    \includegraphics[width=1.0\linewidth]{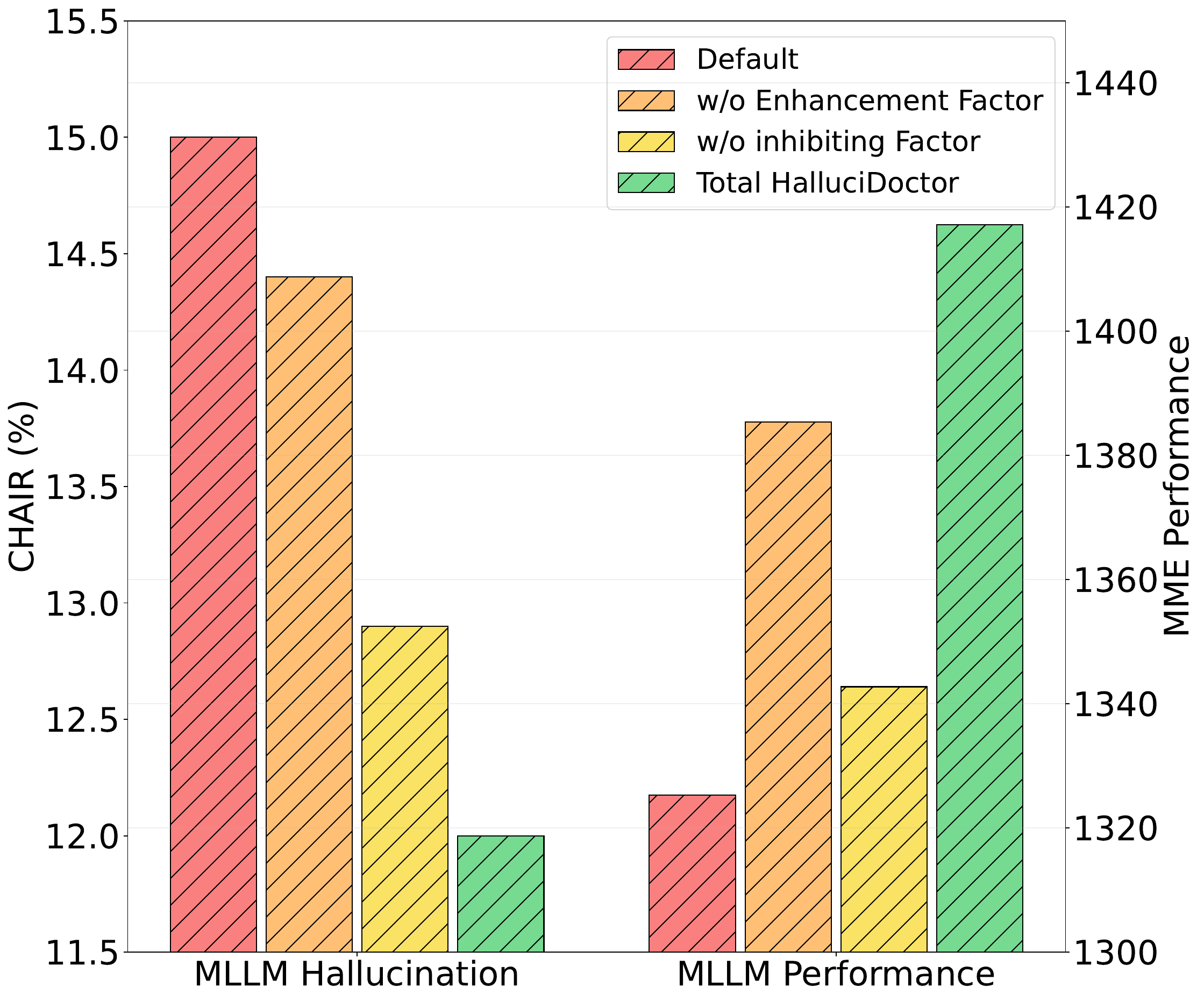}
    \caption{mPLUG-Owl}
  \end{subfigure}
  \caption{Ablation study of two factors in HalluciDoctor.}
  \label{ablation_factor}
  \vspace{-6mm}
\end{figure}

\noindent\textbf{Robustness Analysis of HalluciDoctor.} Additionally, we investigated the robustness of HalluciDoctor when applied to other machine-generated datasets, such as MiniGPT4-Instruction~\cite{zhu2023minigpt}. As illustrated in Table~\ref{minigpt4_results}, HalluciDoctor similarly reduces hallucination frequency and enhances model performance, demonstrating our method's robustness across various machine-generated visual instructions.
\noindent\textbf{Analysis of Instruction Expansion Factors.} 
To dissect the influence of various factors in our counterfactual instruction expansion, we incrementally removed them and presented the associated results in Figure~\ref{ablation_factor}. The absence of enhancement factors for balancing tail co-occurrences leads MLLMs to more hallucinations, while the lack of inhibitory factors leads to excessive unreasonable instructions, diminishing MLLM performance in contextual understanding.

\noindent\textbf{Visualization Results.} In Figure~\ref{visualization}, we present a case where MLLM progressively enhances response quality after HalluciDoctor mitigates the hallucinatory toxicity from visual instruction data. In this case, HalluciDoctor effectively reduces hallucinatory toxicity introduced by LLaVA (\textit{e.g.}, pedestrian crossing the
street). Furthermore, with the aid of the more robust LLaVA++, MLLMs reduce the impact of specious correlations and enhance the perception of fine-grained~(\textit{e.g.}, shop's name) and unusual content~(\textit{e.g.}, cow out front of shop).
More case analyses are in Appendix \textcolor{red}{D}.
\begin{table}[t]
\resizebox{0.475\textwidth}{!}{
\begin{tabular}{lcccc}
\hline
\toprule
\multicolumn{1}{c}{\multirow{2}{*}{Dataset}} & \multicolumn{2}{c}{$\overline{\text{CHAIR}}$~(\%)} & \multicolumn{2}{c}{MME Performance} \\
& $\text{CHAIR}_{I}$~$\downarrow$&$\text{CHAIR}_{S}$~$\downarrow$&Perception~$\uparrow$&Cognition~$\uparrow$\\\hline
 MiniGPT4-Instruction~\cite{zhu2023minigpt} &9.2 & 23.7& 616.41  & 232.71\\ 
 MiniGPT4-Instruction+ & 8.4 &18.6& 659.67&255.03\\
 MiniGPT4-Instruction++ & \textbf{5.9} & \textbf{15.2}& \textbf{696.96} & \textbf{282.86} \\
\bottomrule
\end{tabular}
}
\vspace{-1mm}
\caption{CHAIR results and MME performance of applying HalluciDoctor on MiniGPT4-Instruction Dataset~\cite{zhu2023minigpt}. }
\vspace{-2mm}
\label{minigpt4_results}
\end{table}

\section{Conclusions}
In this paper, we initially delve deep into the underlying hallucination phenomena in machine-generated visual instruction data.
We introduce a flexible framework, HalluciDoctor, that exploits a question-based cross-checking paradigm to detect and eliminate potential hallucinations automatically. 
Additionally, we pinpoint the co-occurrence issue leading to hallucinations and augment the MLLMs' resistance to such errors through the expansion of counterfactual instruction.
The extensive experimental results by both automatic metrics and human evaluations demonstrate the superiority of our approach in dispelling various hallucinations and retaining MLLM's open-ended capabilities.

\noindent\textbf{Acknowledgment.} This work was supported by the National Natural Science Foundation of China (U2336212), Key Research and Development Projects in Zhejiang Province (No. 2024C01106), the National Key Research and Development Project of China (2018AAA0101900), and Research funding from FinVolution Group. We thank all the reviewers for their valuable comments.

\clearpage
{
    \small
    \bibliographystyle{ieeenat_fullname}
    \bibliography{main}

\begin{thebibliography}{54}
\providecommand{\natexlab}[1]{#1}
\providecommand{\url}[1]{\texttt{#1}}
\expandafter\ifx\csname urlstyle\endcsname\relax
  \providecommand{\doi}[1]{doi: #1}\else
  \providecommand{\doi}{doi: \begingroup \urlstyle{rm}\Url}\fi

\bibitem[Agrawal et~al.(2019)Agrawal, Desai, Wang, Chen, Jain, Johnson, Batra, Parikh, Lee, and Anderson]{agrawal2019nocaps}
Harsh Agrawal, Karan Desai, Yufei Wang, Xinlei Chen, Rishabh Jain, Mark Johnson, Dhruv Batra, Devi Parikh, Stefan Lee, and Peter Anderson.
\newblock Nocaps: Novel object captioning at scale.
\newblock In \emph{Proceedings of the IEEE/CVF international conference on computer vision}, pages 8948--8957, 2019.

\bibitem[Alayrac et~al.(2022)Alayrac, Donahue, Luc, Miech, Barr, Hasson, Lenc, Mensch, Millican, Reynolds, et~al.]{alayrac2022flamingo}
Jean-Baptiste Alayrac, Jeff Donahue, Pauline Luc, Antoine Miech, Iain Barr, Yana Hasson, Karel Lenc, Arthur Mensch, Katherine Millican, Malcolm Reynolds, et~al.
\newblock Flamingo: a visual language model for few-shot learning.
\newblock \emph{Advances in Neural Information Processing Systems}, 35:\penalty0 23716--23736, 2022.

\bibitem[Bai et~al.(2023)Bai, Bai, Yang, Wang, Tan, Wang, Lin, Zhou, and Zhou]{bai2023qwen}
Jinze Bai, Shuai Bai, Shusheng Yang, Shijie Wang, Sinan Tan, Peng Wang, Junyang Lin, Chang Zhou, and Jingren Zhou.
\newblock Qwen-vl: A frontier large vision-language model with versatile abilities.
\newblock \emph{arXiv preprint arXiv:2308.12966}, 2023.

\bibitem[Bulian et~al.(2022)Bulian, Buck, Gajewski, Boerschinger, and Schuster]{bulian2022tomayto}
Jannis Bulian, Christian Buck, Wojciech Gajewski, Benjamin Boerschinger, and Tal Schuster.
\newblock Tomayto, tomahto. beyond token-level answer equivalence for question answering evaluation.
\newblock \emph{arXiv preprint arXiv:2202.07654}, 2022.

\bibitem[Chiang et~al.(2023)Chiang, Li, Lin, Sheng, Wu, Zhang, Zheng, Zhuang, Zhuang, Gonzalez, et~al.]{chiang2023vicuna}
Wei-Lin Chiang, Zhuohan Li, Zi Lin, Ying Sheng, Zhanghao Wu, Hao Zhang, Lianmin Zheng, Siyuan Zhuang, Yonghao Zhuang, Joseph~E Gonzalez, et~al.
\newblock Vicuna: An open-source chatbot impressing gpt-4 with 90\%* chatgpt quality.
\newblock \emph{See https://vicuna. lmsys. org (accessed 14 April 2023)}, 2023.

\bibitem[Dai et~al.(2023)Dai, Li, Li, Tiong, Zhao, Wang, Li, Fung, and Hoi]{dai2023instructblip}
Wenliang Dai, Junnan Li, Dongxu Li, Anthony Meng~Huat Tiong, Junqi Zhao, Weisheng Wang, Boyang Li, Pascale Fung, and Steven Hoi.
\newblock Instructblip: Towards general-purpose vision-language models with instruction tuning, 2023.

\bibitem[Feng et~al.(2023)Feng, Balachandran, Bai, and Tsvetkov]{feng2023factkb}
Shangbin Feng, Vidhisha Balachandran, Yuyang Bai, and Yulia Tsvetkov.
\newblock Factkb: Generalizable factuality evaluation using language models enhanced with factual knowledge.
\newblock \emph{arXiv preprint arXiv:2305.08281}, 2023.

\bibitem[Fu et~al.(2023)Fu, Chen, Shen, Qin, Zhang, Lin, Qiu, Lin, Yang, Zheng, et~al.]{fu2023mme}
Chaoyou Fu, Peixian Chen, Yunhang Shen, Yulei Qin, Mengdan Zhang, Xu Lin, Zhenyu Qiu, Wei Lin, Jinrui Yang, Xiawu Zheng, et~al.
\newblock Mme: A comprehensive evaluation benchmark for multimodal large language models.
\newblock \emph{arXiv preprint arXiv:2306.13394}, 2023.

\bibitem[Gunjal et~al.(2023)Gunjal, Yin, and Bas]{gunjal2023detecting}
Anisha Gunjal, Jihan Yin, and Erhan Bas.
\newblock Detecting and preventing hallucinations in large vision language models.
\newblock \emph{arXiv preprint arXiv:2308.06394}, 2023.

\bibitem[Guo et~al.(2023)Guo, Li, Li, Tiong, Li, Tao, and Hoi]{guo2023images}
Jiaxian Guo, Junnan Li, Dongxu Li, Anthony Meng~Huat Tiong, Boyang Li, Dacheng Tao, and Steven Hoi.
\newblock From images to textual prompts: Zero-shot visual question answering with frozen large language models.
\newblock In \emph{Proceedings of the IEEE/CVF Conference on Computer Vision and Pattern Recognition}, pages 10867--10877, 2023.

\bibitem[Hu et~al.(2021)Hu, Shen, Wallis, Allen-Zhu, Li, Wang, Wang, and Chen]{hu2021lora}
Edward~J Hu, Yelong Shen, Phillip Wallis, Zeyuan Allen-Zhu, Yuanzhi Li, Shean Wang, Lu Wang, and Weizhu Chen.
\newblock Lora: Low-rank adaptation of large language models.
\newblock \emph{arXiv preprint arXiv:2106.09685}, 2021.

\bibitem[Hudson and Manning(2019)]{hudson2019gqa}
Drew~A Hudson and Christopher~D Manning.
\newblock Gqa: A new dataset for real-world visual reasoning and compositional question answering.
\newblock In \emph{Proceedings of the IEEE/CVF conference on computer vision and pattern recognition}, pages 6700--6709, 2019.

\bibitem[Krishna et~al.(2017)Krishna, Zhu, Groth, Johnson, Hata, Kravitz, Chen, Kalantidis, Li, Shamma, et~al.]{krishna2017visual}
Ranjay Krishna, Yuke Zhu, Oliver Groth, Justin Johnson, Kenji Hata, Joshua Kravitz, Stephanie Chen, Yannis Kalantidis, Li-Jia Li, David~A Shamma, et~al.
\newblock Visual genome: Connecting language and vision using crowdsourced dense image annotations.
\newblock \emph{International journal of computer vision}, 123\penalty0 (1):\penalty0 32--73, 2017.

\bibitem[Le et~al.(2023)Le, Lal, and Howard]{le2023coco}
Tiep Le, Vasudev Lal, and Phillip Howard.
\newblock Coco-counterfactuals: Automatically constructed counterfactual examples for image-text pairs.
\newblock \emph{arXiv preprint arXiv:2309.14356}, 2023.

\bibitem[Li et~al.(2023{\natexlab{a}})Li, Zhang, Chen, Wang, Yang, and Liu]{li2023otter}
Bo Li, Yuanhan Zhang, Liangyu Chen, Jinghao Wang, Jingkang Yang, and Ziwei Liu.
\newblock Otter: A multi-modal model with in-context instruction tuning.
\newblock \emph{arXiv preprint arXiv:2305.03726}, 2023{\natexlab{a}}.

\bibitem[Li et~al.(2022)Li, Li, Xiong, and Hoi]{li2022blip}
Junnan Li, Dongxu Li, Caiming Xiong, and Steven Hoi.
\newblock Blip: Bootstrapping language-image pre-training for unified vision-language understanding and generation.
\newblock In \emph{International Conference on Machine Learning}, pages 12888--12900. PMLR, 2022.

\bibitem[Li et~al.(2023{\natexlab{b}})Li, Li, Savarese, and Hoi]{li2023blip}
Junnan Li, Dongxu Li, Silvio Savarese, and Steven Hoi.
\newblock Blip-2: Bootstrapping language-image pre-training with frozen image encoders and large language models.
\newblock \emph{arXiv preprint arXiv:2301.12597}, 2023{\natexlab{b}}.

\bibitem[Li et~al.(2023{\natexlab{c}})Li, Pan, Ge, Gao, Zhang, Ji, Zhang, Chua, Tang, and Zhuang]{li2023fine}
Juncheng Li, Kaihang Pan, Zhiqi Ge, Minghe Gao, Hanwang Zhang, Wei Ji, Wenqiao Zhang, Tat-Seng Chua, Siliang Tang, and Yueting Zhuang.
\newblock Fine-tuning multimodal llms to follow zero-shot demonstrative instructions.
\newblock In \emph{The Twelfth International Conference on Learning Representations}, 2023{\natexlab{c}}.

\bibitem[Li et~al.(2023{\natexlab{d}})Li, Du, Zhou, Wang, Zhao, and Wen]{li2023evaluating}
Yifan Li, Yifan Du, Kun Zhou, Jinpeng Wang, Wayne~Xin Zhao, and Ji-Rong Wen.
\newblock Evaluating object hallucination in large vision-language models.
\newblock \emph{arXiv preprint arXiv:2305.10355}, 2023{\natexlab{d}}.

\bibitem[Li et~al.(2023{\natexlab{e}})Li, Xu, Miao, Zhou, and Qian]{li2023large}
Yongqi Li, Mayi Xu, Xin Miao, Shen Zhou, and Tieyun Qian.
\newblock Large language models as counterfactual generator: Strengths and weaknesses.
\newblock \emph{arXiv preprint arXiv:2305.14791}, 2023{\natexlab{e}}.

\bibitem[Li et~al.(2023{\natexlab{f}})Li, Chai, Yue, Qu, Haffari, Li, Ji, and Tran]{li2023factual}
Zhuang Li, Yuyang Chai, Terry~Zhuo Yue, Lizhen Qu, Gholamreza Haffari, Fei Li, Donghong Ji, and Quan~Hung Tran.
\newblock Factual: A benchmark for faithful and consistent textual scene graph parsing.
\newblock \emph{arXiv preprint arXiv:2305.17497}, 2023{\natexlab{f}}.

\bibitem[Liao et~al.(2023)Liao, Zhang, Xia, Zhang, Cao, Qiao, and Yan]{liao2023revo}
Ning Liao, Shaofeng Zhang, Renqiu Xia, Bo Zhang, Min Cao, Yu Qiao, and Junchi Yan.
\newblock Revo-lion: Evaluating and refining vision-language instruction tuning datasets.
\newblock \emph{arXiv preprint arXiv:2310.06594}, 2023.

\bibitem[Lin et~al.(2014)Lin, Maire, Belongie, Hays, Perona, Ramanan, Doll{\'a}r, and Zitnick]{lin2014microsoft}
Tsung-Yi Lin, Michael Maire, Serge Belongie, James Hays, Pietro Perona, Deva Ramanan, Piotr Doll{\'a}r, and C~Lawrence Zitnick.
\newblock Microsoft coco: Common objects in context.
\newblock In \emph{European conference on computer vision}, pages 740--755. Springer, 2014.

\bibitem[Liu et~al.(2023{\natexlab{a}})Liu, Lin, Li, Wang, Yacoob, and Wang]{liu2023aligning}
Fuxiao Liu, Kevin Lin, Linjie Li, Jianfeng Wang, Yaser Yacoob, and Lijuan Wang.
\newblock Aligning large multi-modal model with robust instruction tuning.
\newblock \emph{arXiv preprint arXiv:2306.14565}, 2023{\natexlab{a}}.

\bibitem[Liu et~al.(2023{\natexlab{b}})Liu, Li, Wu, and Lee]{liu2023visual}
Haotian Liu, Chunyuan Li, Qingyang Wu, and Yong~Jae Lee.
\newblock Visual instruction tuning.
\newblock \emph{arXiv preprint arXiv:2304.08485}, 2023{\natexlab{b}}.

\bibitem[Liu et~al.(2023{\natexlab{c}})Liu, Zeng, Ren, Li, Zhang, Yang, Li, Yang, Su, Zhu, et~al.]{liu2023grounding}
Shilong Liu, Zhaoyang Zeng, Tianhe Ren, Feng Li, Hao Zhang, Jie Yang, Chunyuan Li, Jianwei Yang, Hang Su, Jun Zhu, et~al.
\newblock Grounding dino: Marrying dino with grounded pre-training for open-set object detection.
\newblock \emph{arXiv preprint arXiv:2303.05499}, 2023{\natexlab{c}}.

\bibitem[Lu et~al.(2023)Lu, Rao, Chen, Guo, Zhang, Sun, Yang, and Yang]{lu2023evaluation}
Jiaying Lu, Jinmeng Rao, Kezhen Chen, Xiaoyuan Guo, Yawen Zhang, Baochen Sun, Carl Yang, and Jie Yang.
\newblock Evaluation and mitigation of agnosia in multimodal large language models.
\newblock \emph{arXiv preprint arXiv:2309.04041}, 2023.

\bibitem[Marino et~al.(2019)Marino, Rastegari, Farhadi, and Mottaghi]{marino2019ok}
Kenneth Marino, Mohammad Rastegari, Ali Farhadi, and Roozbeh Mottaghi.
\newblock Ok-vqa: A visual question answering benchmark requiring external knowledge.
\newblock In \emph{Proceedings of the IEEE/cvf conference on computer vision and pattern recognition}, pages 3195--3204, 2019.

\bibitem[OpenAI(2023{\natexlab{a}})]{chatgpt}
OpenAI.
\newblock {ChatGPT.}
\newblock \url{https://openai.com/blog/chatgpt/}, 2023{\natexlab{a}}.

\bibitem[OpenAI(2023{\natexlab{b}})]{gpt4}
OpenAI.
\newblock {GPT-4.}
\newblock \url{https://openai.com/gpt-4}, 2023{\natexlab{b}}.

\bibitem[Ouyang et~al.(2022)Ouyang, Wu, Jiang, Almeida, Wainwright, Mishkin, Zhang, Agarwal, Slama, Ray, et~al.]{ouyang2022training}
Long Ouyang, Jeffrey Wu, Xu Jiang, Diogo Almeida, Carroll Wainwright, Pamela Mishkin, Chong Zhang, Sandhini Agarwal, Katarina Slama, Alex Ray, et~al.
\newblock Training language models to follow instructions with human feedback.
\newblock \emph{Advances in Neural Information Processing Systems}, 35:\penalty0 27730--27744, 2022.

\bibitem[Pawlowski et~al.(2020)Pawlowski, Coelho~de Castro, and Glocker]{pawlowski2020deep}
Nick Pawlowski, Daniel Coelho~de Castro, and Ben Glocker.
\newblock Deep structural causal models for tractable counterfactual inference.
\newblock \emph{Advances in neural information processing systems}, 33:\penalty0 857--869, 2020.

\bibitem[Peng et~al.(2023)Peng, Li, He, Galley, and Gao]{peng2023instruction}
Baolin Peng, Chunyuan Li, Pengcheng He, Michel Galley, and Jianfeng Gao.
\newblock Instruction tuning with gpt-4.
\newblock \emph{arXiv preprint arXiv:2304.03277}, 2023.

\bibitem[Rohrbach et~al.(2018)Rohrbach, Hendricks, Burns, Darrell, and Saenko]{rohrbach2018object}
Anna Rohrbach, Lisa~Anne Hendricks, Kaylee Burns, Trevor Darrell, and Kate Saenko.
\newblock Object hallucination in image captioning.
\newblock \emph{arXiv preprint arXiv:1809.02156}, 2018.

\bibitem[Rombach et~al.(2022)Rombach, Blattmann, Lorenz, Esser, and Ommer]{rombach2022high}
Robin Rombach, Andreas Blattmann, Dominik Lorenz, Patrick Esser, and Bj{\"o}rn Ommer.
\newblock High-resolution image synthesis with latent diffusion models.
\newblock In \emph{Proceedings of the IEEE/CVF Conference on Computer Vision and Pattern Recognition}, pages 10684--10695, 2022.

\bibitem[Touvron et~al.(2023)Touvron, Martin, Stone, Albert, Almahairi, Babaei, Bashlykov, Batra, Bhargava, Bhosale, et~al.]{touvron2023llama}
Hugo Touvron, Louis Martin, Kevin Stone, Peter Albert, Amjad Almahairi, Yasmine Babaei, Nikolay Bashlykov, Soumya Batra, Prajjwal Bhargava, Shruti Bhosale, et~al.
\newblock Llama 2: Open foundation and fine-tuned chat models.
\newblock \emph{arXiv preprint arXiv:2307.09288}, 2023.

\bibitem[Tsimpoukelli et~al.(2021)Tsimpoukelli, Menick, Cabi, Eslami, Vinyals, and Hill]{tsimpoukelli2021multimodal}
Maria Tsimpoukelli, Jacob~L Menick, Serkan Cabi, SM Eslami, Oriol Vinyals, and Felix Hill.
\newblock Multimodal few-shot learning with frozen language models.
\newblock \emph{Advances in Neural Information Processing Systems}, 34:\penalty0 200--212, 2021.

\bibitem[Wang et~al.(2023)Wang, Wu, Han, Peng, Zhong, Zhang, Dong, Li, Li, Wang, et~al.]{wang2023vigc}
Bin Wang, Fan Wu, Xiao Han, Jiahui Peng, Huaping Zhong, Pan Zhang, Xiaoyi Dong, Weijia Li, Wei Li, Jiaqi Wang, et~al.
\newblock Vigc: Visual instruction generation and correction.
\newblock \emph{arXiv preprint arXiv:2308.12714}, 2023.

\bibitem[Wang et~al.(2022)Wang, Kordi, Mishra, Liu, Smith, Khashabi, and Hajishirzi]{wang2022self}
Yizhong Wang, Yeganeh Kordi, Swaroop Mishra, Alisa Liu, Noah~A Smith, Daniel Khashabi, and Hannaneh Hajishirzi.
\newblock Self-instruct: Aligning language model with self generated instructions.
\newblock \emph{arXiv preprint arXiv:2212.10560}, 2022.

\bibitem[Wei et~al.(2022)Wei, Wang, Schuurmans, Bosma, Xia, Chi, Le, Zhou, et~al.]{wei2022chain}
Jason Wei, Xuezhi Wang, Dale Schuurmans, Maarten Bosma, Fei Xia, Ed Chi, Quoc~V Le, Denny Zhou, et~al.
\newblock Chain-of-thought prompting elicits reasoning in large language models.
\newblock \emph{Advances in Neural Information Processing Systems}, 35:\penalty0 24824--24837, 2022.

\bibitem[Wu et~al.(2023)Wu, Qiu, Ross, Aky{\"u}rek, Chen, Wang, Kim, Andreas, and Kim]{wu2023reasoning}
Zhaofeng Wu, Linlu Qiu, Alexis Ross, Ekin Aky{\"u}rek, Boyuan Chen, Bailin Wang, Najoung Kim, Jacob Andreas, and Yoon Kim.
\newblock Reasoning or reciting? exploring the capabilities and limitations of language models through counterfactual tasks.
\newblock \emph{arXiv preprint arXiv:2307.02477}, 2023.

\bibitem[Ye et~al.(2023)Ye, Xu, Xu, Ye, Yan, Zhou, Wang, Hu, Shi, Shi, et~al.]{ye2023mplug}
Qinghao Ye, Haiyang Xu, Guohai Xu, Jiabo Ye, Ming Yan, Yiyang Zhou, Junyang Wang, Anwen Hu, Pengcheng Shi, Yaya Shi, et~al.
\newblock mplug-owl: Modularization empowers large language models with multimodality.
\newblock \emph{arXiv preprint arXiv:2304.14178}, 2023.

\bibitem[Yin et~al.(2023{\natexlab{a}})Yin, Fu, Zhao, Li, Sun, Xu, and Chen]{yin2023survey}
Shukang Yin, Chaoyou Fu, Sirui Zhao, Ke Li, Xing Sun, Tong Xu, and Enhong Chen.
\newblock A survey on multimodal large language models.
\newblock \emph{arXiv preprint arXiv:2306.13549}, 2023{\natexlab{a}}.

\bibitem[Yin et~al.(2023{\natexlab{b}})Yin, Fu, Zhao, Xu, Wang, Sui, Shen, Li, Sun, and Chen]{yin2023woodpecker}
Shukang Yin, Chaoyou Fu, Sirui Zhao, Tong Xu, Hao Wang, Dianbo Sui, Yunhang Shen, Ke Li, Xing Sun, and Enhong Chen.
\newblock Woodpecker: Hallucination correction for multimodal large language models.
\newblock \emph{arXiv preprint arXiv:2310.16045}, 2023{\natexlab{b}}.

\bibitem[Yu et~al.(2023{\natexlab{a}})Yu, Li, Wu, Tang, Ji, and Zhuang]{yu2023visually}
Qifan Yu, Juncheng Li, Yu Wu, Siliang Tang, Wei Ji, and Yueting Zhuang.
\newblock Visually-prompted language model for fine-grained scene graph generation in an open world.
\newblock In \emph{Proceedings of the IEEE/CVF International Conference on Computer Vision}, pages 21560--21571, 2023{\natexlab{a}}.

\bibitem[Yu et~al.(2023{\natexlab{b}})Yu, Li, Ye, Tang, and Zhuang]{yu2023interactive}
Qifan Yu, Juncheng Li, Wentao Ye, Siliang Tang, and Yueting Zhuang.
\newblock Interactive data synthesis for systematic vision adaptation via llms-aigcs collaboration.
\newblock \emph{arXiv preprint arXiv:2305.12799}, 2023{\natexlab{b}}.

\bibitem[Zha et~al.(2023)Zha, Yang, Li, and Hu]{zha2023alignscore}
Yuheng Zha, Yichi Yang, Ruichen Li, and Zhiting Hu.
\newblock Alignscore: Evaluating factual consistency with a unified alignment function.
\newblock \emph{arXiv preprint arXiv:2305.16739}, 2023.

\bibitem[Zhang et~al.(2022{\natexlab{a}})Zhang, Ma, Wang, and Chua]{zhang2022incorporating}
An Zhang, Wenchang Ma, Xiang Wang, and Tat-Seng Chua.
\newblock Incorporating bias-aware margins into contrastive loss for collaborative filtering.
\newblock \emph{Advances in Neural Information Processing Systems}, 35:\penalty0 7866--7878, 2022{\natexlab{a}}.

\bibitem[Zhang et~al.(2023{\natexlab{a}})Zhang, Zhai, Zhao, Wen, and Zhao]{zhang2023if}
Letian Zhang, Xiaotong Zhai, Zhongkai Zhao, Xin Wen, and Bingchen Zhao.
\newblock What if the tv was off? examining counterfactual reasoning abilities of multi-modal language models.
\newblock In \emph{Proceedings of the IEEE/CVF International Conference on Computer Vision}, pages 4629--4633, 2023{\natexlab{a}}.

\bibitem[Zhang et~al.(2022{\natexlab{b}})Zhang, Shi, Guo, Zhang, Cai, Li, Luo, and Zhuang]{zhang2022magic}
Wenqiao Zhang, Haochen Shi, Jiannan Guo, Shengyu Zhang, Qingpeng Cai, Juncheng Li, Sihui Luo, and Yueting Zhuang.
\newblock Magic: Multimodal relational graph adversarial inference for diverse and unpaired text-based image captioning.
\newblock In \emph{Proceedings of the AAAI Conference on Artificial Intelligence}, pages 3335--3343, 2022{\natexlab{b}}.

\bibitem[Zhang et~al.(2023{\natexlab{b}})Zhang, Liu, Zeng, Ooi, Tang, and Zhuang]{zhang2023learning}
Wenqiao Zhang, Changshuo Liu, Lingze Zeng, Bengchin Ooi, Siliang Tang, and Yueting Zhuang.
\newblock Learning in imperfect environment: Multi-label classification with long-tailed distribution and partial labels.
\newblock In \emph{Proceedings of the IEEE/CVF International Conference on Computer Vision}, pages 1423--1432, 2023{\natexlab{b}}.

\bibitem[Zhao et~al.(2023)Zhao, Zhou, Li, Tang, Wang, Hou, Min, Zhang, Zhang, Dong, et~al.]{zhao2023survey}
Wayne~Xin Zhao, Kun Zhou, Junyi Li, Tianyi Tang, Xiaolei Wang, Yupeng Hou, Yingqian Min, Beichen Zhang, Junjie Zhang, Zican Dong, et~al.
\newblock A survey of large language models.
\newblock \emph{arXiv preprint arXiv:2303.18223}, 2023.

\bibitem[Zhou et~al.(2023)Zhou, Cui, Yoon, Zhang, Deng, Finn, Bansal, and Yao]{zhou2023analyzing}
Yiyang Zhou, Chenhang Cui, Jaehong Yoon, Linjun Zhang, Zhun Deng, Chelsea Finn, Mohit Bansal, and Huaxiu Yao.
\newblock Analyzing and mitigating object hallucination in large vision-language models.
\newblock \emph{arXiv preprint arXiv:2310.00754}, 2023.

\bibitem[Zhu et~al.(2023)Zhu, Chen, Shen, Li, and Elhoseiny]{zhu2023minigpt}
Deyao Zhu, Jun Chen, Xiaoqian Shen, Xiang Li, and Mohamed Elhoseiny.
\newblock Minigpt-4: Enhancing vision-language understanding with advanced large language models.
\newblock \emph{arXiv preprint arXiv:2304.10592}, 2023.

\end{thebibliography}
}
\clearpage
\setcounter{page}{1}
\maketitlesupplementary

\appendix
\section{Overview}
In this supplementary material, we present:

\begin{itemize}
	\item More detailed analysis of HalluciDoctor~(Section \ref{2}).
	
	\item More experimental analysis~(Section \ref{3}).
	
	\item Additional examples~(Section \ref{4}).

\end{itemize}
\section{HalluciDoctor Framework}
\label{2}
\subsection{Question Generation} 
This section provides specific steps for answer-based question generation. We utilize ChatGPT~\cite{chatgpt} as the powerful question generator, covering a broad spectrum of semantic chunks and various question types. Specifically, we construct the prompt template shown in Figure~\ref{qg_prompt}~(a), filling the context description and answer blocks into slots to generate all corresponding questions. These questions, covering various types, effectively reflect meaningful semantic information in the descriptions.
    
\subsection{Consistency Cross-checking Analysis}  
Considering the importance of the threshold in consistency cross-checking for identifying hallucinatory chunks, we further explore the effect of consistency cross-checking under different threshold setups in Table~\ref{consistency_threshold}. Here, we assess the impact of different consistency thresholds by evaluating the average performance on CHAIR and MME benchmarks. 

Our observations indicate that at lower consistency thresholds, as the threshold increases, hallucinatory descriptions are progressively detected and eliminated. This provides MLLMs with higher-quality instruction data for fine-tuning, thereby reducing the likelihood of hallucinatory outputs while enhancing model performance. However, when the consistency threshold exceeds 0.5, there's a significant decline in model performance. The possible reason is that HalluciDoctor eliminates almost all answer blocks as hallucinations when the consistency threshold is too high, resulting in a substantial loss of accurate semantics. Therefore, to effectively reduce hallucinations in MLLM outputs while ensuring competitive performance, we select \textbf{0.5} as the final threshold for the consistency cross-checking stage in our HalluciDoctor framework. To enhance the precision of hallucination detection and elimination, we will explore more advanced approaches for computing consistency scores~\cite{feng2023factkb, zha2023alignscore} in the future.
\begin{table}[ht]
\renewcommand\arraystretch{1.5}
\resizebox{0.475\textwidth}{!}{
\begin{tabular}{c|c|ccccc}
\toprule
 Consistency threshold & w/o HalluciDoctor & 0.1 & 0.3 & 0.5 & 0.7 & 0.9  \\ \hline
$\overline{\text{CHAIR}}$~(\%)~$\downarrow$ & 21.7 & 19.9 & 16.1 & \underline{13.8} & 14.2 & \textbf{13.6} \\ \hline
$\overline{\text{MME performance}}$~(\%)~$\uparrow$ &  1148.9 &  1153.2 &  \underline{1178.6} & \textbf{1207.2} &  1012.9 &  739.1 \\ 
\bottomrule
\end{tabular}
}
\caption{The influence of different consistency threshold for the hallucination elimination in visual instruction data.}
\label{consistency_threshold}
\end{table}
\subsection{Hallucination Elimination}
\begin{figure*}
  \centering
  \begin{subfigure}{1.0\linewidth}
    \includegraphics[width=1.\linewidth]{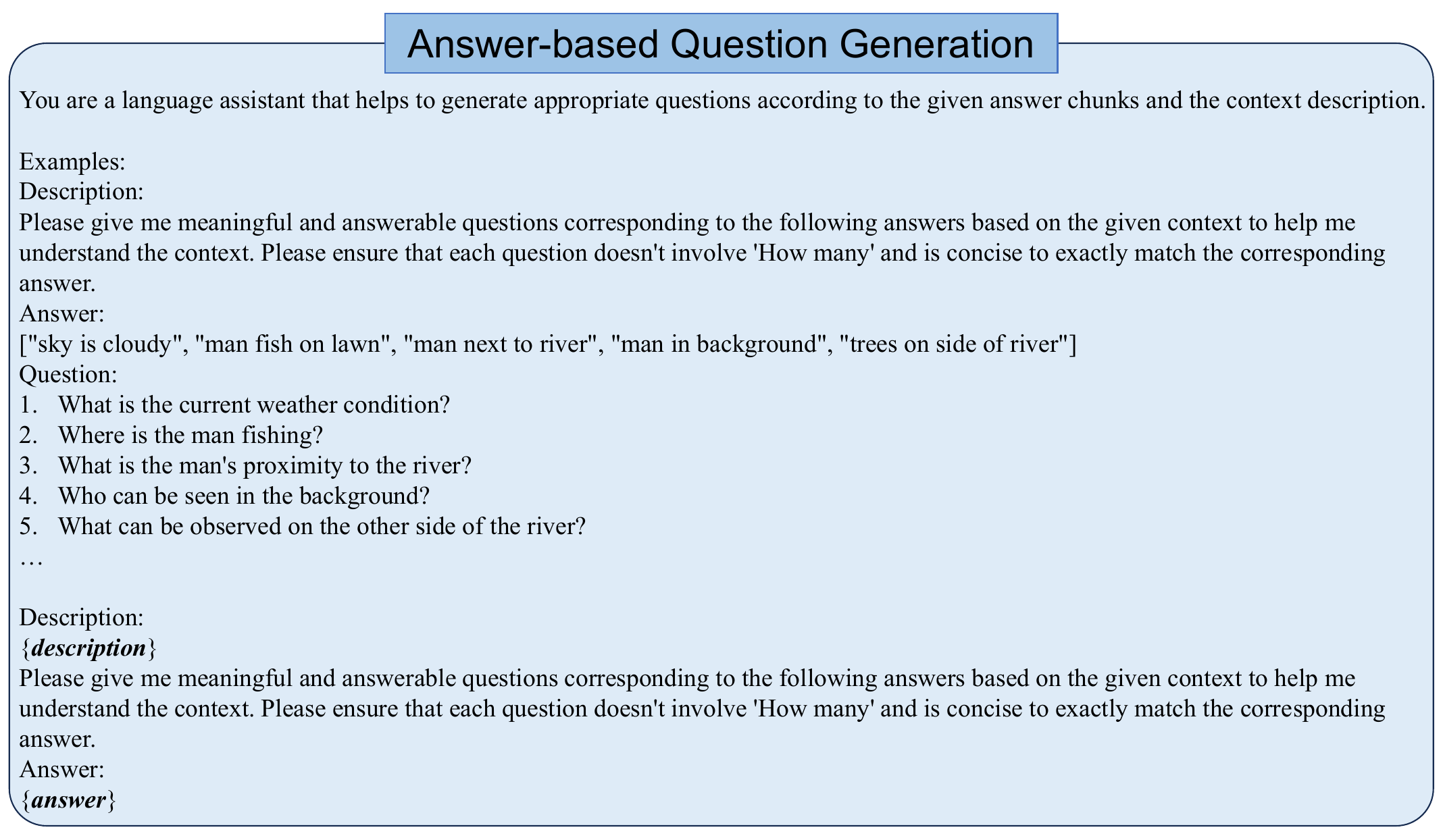}
  \caption{The details of the prompt design for \textit{Question Generation} in HalluciDoctor.}
  \end{subfigure}
  \begin{subfigure}{1.0\linewidth}
    \includegraphics[width=1.\linewidth]{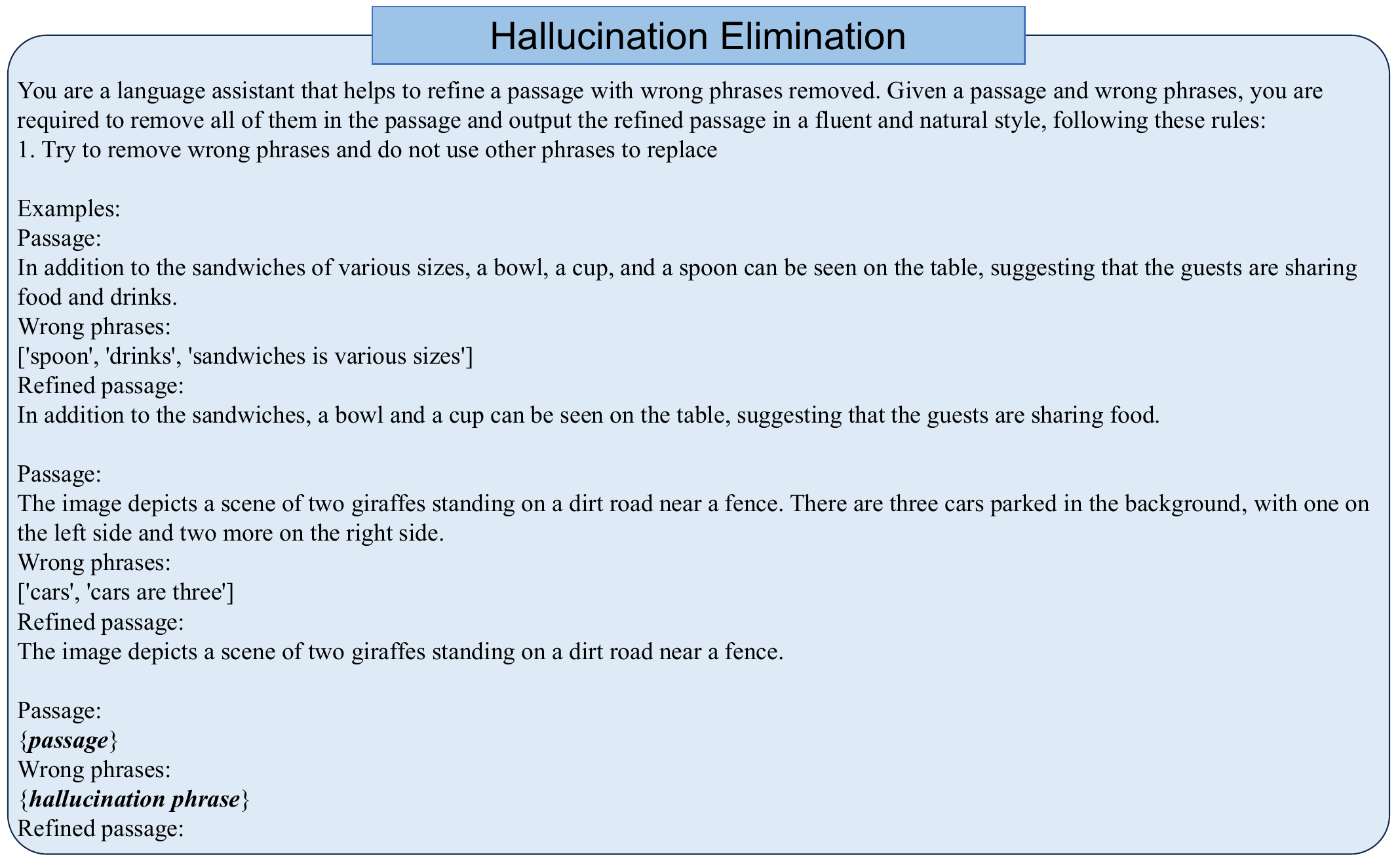}
  \caption{The details of the prompt design for \textit{Hallucination Elimination} in HalluciDoctor.}
  \end{subfigure}
  \caption{The details of the prompt design in HalluciDoctor. There are injectable slots in the prompts, such as \textbf{\textit{description}}, \textbf{\textit{answer}}, \textbf{\textit{passage}}, and \textbf{\textit{hallucination phrase}}. These slots are uniformly replaced with the corresponding text before being fed into the LLM.}
  \label{qg_prompt}
\end{figure*}

This section provides specific steps for hallucination elimination in the visual instruction data. To eliminate detected hallucinations while preserving meaningful semantics in the original descriptions, we employ ChatGPT to refine the descriptions. Specifically, we input both the hallucinatory phrases and the original descriptions into the prompt template shown in Figure~\ref{qg_prompt}~(b), prompting ChatGPT to remove hallucinatory phrases without altering the sentence structure. We also show some examples in Figure~\ref{qg_prompt}~(b). These refined descriptions are then employed to update the visual instructions in LLaVA, efficiently creating the rectified dataset LLaVA++.

\subsection{Visual Instruction Expansion}
This section provides detailed steps for selecting target images and how to add hallucinatory objects into target scenes.
Firstly, we filter target scenes based on object detection to ensure the specific hallucinatory object $o$ is absent. We then generate candidate objects and their corresponding masks for counterfactual image synthesis using text-to-image models~\cite{rombach2022high, yu2023interactive} and object detection tools~\cite{liu2023grounding}. Subsequently, we provide the LLM with image sizes and foreground object locations of target scenes to enable it to determine suitable positions and scaling. Finally, we employ structure-preserving filtering based on the depth map L1 distance for natural image incorporation.

In this way, counterfactual instruction expansion focuses on detailed and unusual instruction modifications, necessitating MLLMs to perceive fine-grained concepts for comprehensive instruction alignment. Consequently, it will generate fewer hallucinations~(\textit{e.g.}, 13.8\% $\rightarrow$ 12.0\% in MiniGPT-4) and demonstrate superior proficiency in perceiving specific information~(\textit{e.g.}, shop’s name). This approach can also alleviate the adverse impact of long-tail distributions in various domains~\cite{zhang2023learning}.
\begin{table*}
    \centering
    \begin{subtable}[t]{0.33\linewidth}
        \begin{tabular}{lcc}
            \toprule
            Dataset & Accuracy & F1 \\
            \midrule
            w/ LLaVA~\cite{liu2023visual} & 75.1 & 77.8\\
            w/ LRV~\cite{liu2023aligning} & 64.4&72.6 \\
            w/ LLaVA+ & 79.1 & 80.0 \\
            w/ LLaVA++ & \textbf{80.1}&\textbf{80.4} \\
            \bottomrule
        \end{tabular}
        \caption{Random Setting}
    \end{subtable}
    \begin{subtable}[t]{0.33\linewidth}
        \begin{tabular}{lcc}
            \toprule
            Dataset & Accuracy & F1 \\
            \midrule
            w/ LLaVA~\cite{liu2023visual} &65.6 &71.7 \\
            w/ LRV~\cite{liu2023aligning} & 63.4&72.3 \\ 
            w/ LLaVA+ & 74.0&75.3 \\
            w/ LLaVA++ & \textbf{76.3}&\textbf{75.9} \\
            \bottomrule
        \end{tabular}
        \caption{Popular Setting}
    \end{subtable}
    \begin{subtable}[t]{0.33\linewidth}
        \begin{tabular}{lcc}
            \toprule
            Dataset & Accuracy & F1 \\
            \midrule
            w/ LLaVA~\cite{liu2023visual} &63.2 &70.5 \\
            w/ LRV~\cite{liu2023aligning} & 60.7&70.6 \\
            w/ LLaVA+ & 68.5&72.0 \\
            w/ LLaVA++ & \textbf{71.9}&\textbf{74.2} \\
            \bottomrule
        \end{tabular}
        \caption{Adversarial Setting}
    \end{subtable}
    \caption{Zero-shot object hallucination results for MiniGPT-4~\cite{zhu2023minigpt} fine-tuned with various visual instructions on POPE~\cite{liu2023aligning} evaluation. We follow the official setup, which involves using three different strategies~(\textsl{i.e.}, random, popular, and adversarial setting) to sample objects not present in the images and then computing the corresponding accuracy and F1 scores.}
    \label{pope_result}
\end{table*}
\section{More Experimental Analysis}
\label{3}
\subsection{Experiment Details} 
\noindent\textbf{Implementation Details.} As for MiniGPT4, we initialize from its checkpoint of the first pretraining stage and only fine-tune the linear projection layer of the model for 10000 steps. As for mPLUG-Owl, we train the text encoder with the LoRA~\cite{hu2021lora} strategy and prepare for 4000 steps. Due to limited computing resources, we set the micro-batch size to 4 and only fine-tuned the 7B model with NVIDIA RTX 3090. To make a fair comparison in our experiments, we only change the visual instruction data under different setups and keep other parameters the same as original models.

\noindent\textbf{Evaluation Setups.} MSCOCO~\cite{lin2014microsoft} is a comprehensive dataset with 80 object categories used for diverse vision tasks. Visual Genome~\cite{krishna2017visual} is another vision dataset with more detailed visual information like bounding boxes and region captions. We select the overlapped images from MSCOCO and VG to construct validation images, aiming to encompass annotations of various objects, relationships, and attributes. Additionally, we employ powerful visual foundation models~\cite{li2023blip, liu2023grounding, zhang2022magic} to identify objects, relations, and attributes of images in the validation set, thereby enriching the ground truth labels. In the validation stage, we will extract object, relation, and attribute phrases from the description of MLLMs that are fine-tuned on different visual instruction datasets, and calculate the corresponding hallucinatory metrics by matching them to ground truth labels.

\begin{figure*}
  \centering
  \includegraphics[width=1.\linewidth]{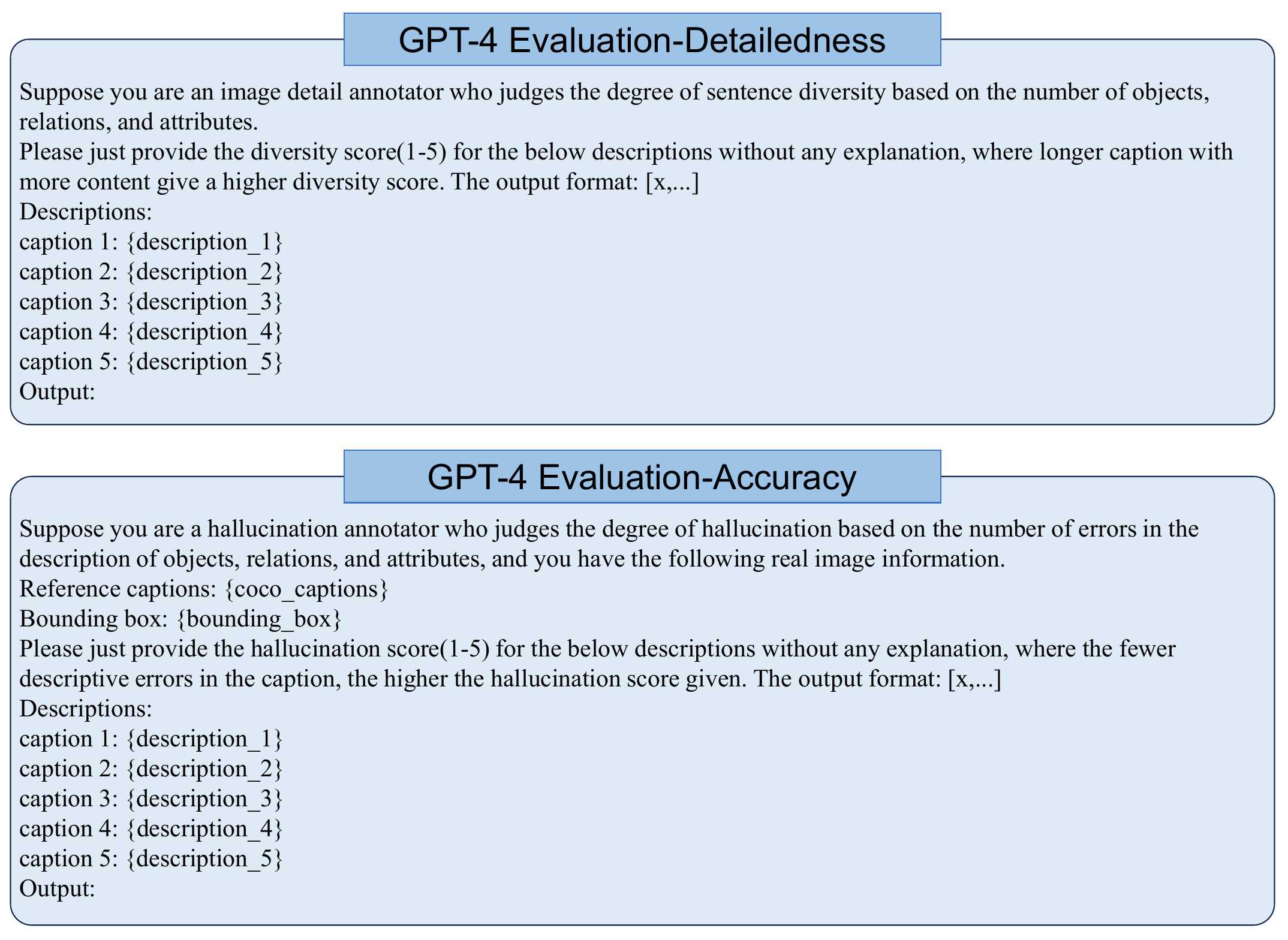}
  \caption{The details of the prompt design for GPT-4 evaluations.}
  \label{gpt4eval_prompt}
\end{figure*}

\noindent\textbf{GPT-4 Evaluation.} We show the GPT-4 evaluation's prompt templates for detailedness and accuracy in Figure~\ref{gpt4eval_prompt}.
\subsection{POPE Results}
We compare the MLLM fine-tuned on our more robust dataset LLaVA+ and LLaVA++ against the baseline dataset on POPE evaluation~\cite{li2023evaluating} in Table~\ref{pope_result}. Although POPE is tailored for close-ended questions of object hallucinations, rendering it unsuitable for our comprehensive evaluation of various hallucinations in visual instruction data, our approach also shows a similar tendency to our main results that LLaVA+ and LLAVA++ from HalluciDoctor achieve consistent gains in all accuracy and F1 score. The results indicate that HalluciDoctor is effective in correcting object-level hallucinations. In addition, the MLLM fine-tuned on LLAVA++ obtains the highest accuracy and F1 score, demonstrating that the more robust visual instruction dataset can enhance MLLMs' ability to discern negative instructions, especially in the more challenging adversarial setting.
\begin{table}[!h]
\vspace{-3.5mm}
\resizebox{0.475\textwidth}{!}{
\begin{tabular}{lccc}
\hline
\toprule
\multicolumn{1}{c}{\multirow{2}{*}{Dataset}} & \textbf{Captioning} & \multicolumn{2}{c}{\textbf{VQA}}\\
&NoCaps~(val)~$\uparrow$&GQA~$\uparrow$&AOK-VQA~$\uparrow$\\\hline
Faithful Prompt &101.5&40.5&56.1\\
LURE~\cite{zhou2023analyzing} & {93.9}&{41.4}&{58.3}\\
VIGC~\cite{wang2023vigc}& {96.6}&{41.0}&{58.9}\\ 
$\text{MiniGPT4}_{\rm +LRV}$~\cite{liu2023aligning} & {103.9}&{40.7}&{57.6}\\ \hline
$\text{MiniGPT4}_{\rm +LLaVA++}$ & 104.1& \textbf{43.7}& 60.1\\
$\text{mPLUG-Owl}_{\rm +LLaVA++}$ & \textbf{104.4}&43.3&\textbf{61.0} \\ 
\bottomrule
\end{tabular}
}
\caption{Overview performance comparison on conventional zero-shot vision-language tasks~(\textsl{i.e.}, captioning, VQA).}
\vspace{-4.5mm}
\label{vqa_results}
\end{table}
\subsection{Zero-shot Vision-Language Task Results}
As a versatile MLLM, the model's performance cannot be compromised by instruction fine-tuning. On the contrary, by eliminating hallucinatory information in the training data, the MLLM demonstrates stronger generalization capabilities for conventional visual tasks. We perform the quantitative evaluation on the zero-shot vision-language tasks based on captioning~(NoCaps~\cite{agrawal2019nocaps}) and visual question answering~(GQA~\cite{hudson2019gqa}, AOK-VQA~\cite{marino2019ok}). Table~\ref{vqa_results} provides an overview of the performance of HalluciDoctor on various zero-shot vision-language tasks. Compared to other works on hallucination elimination, our method achieves better generalization performance on traditional vision tasks.

\section{Additional Examples}
\label{4}
\subsection{Evaluation of Visual Instruction Data}
\begin{figure*}
  \centering
  \includegraphics[width=1.\linewidth]{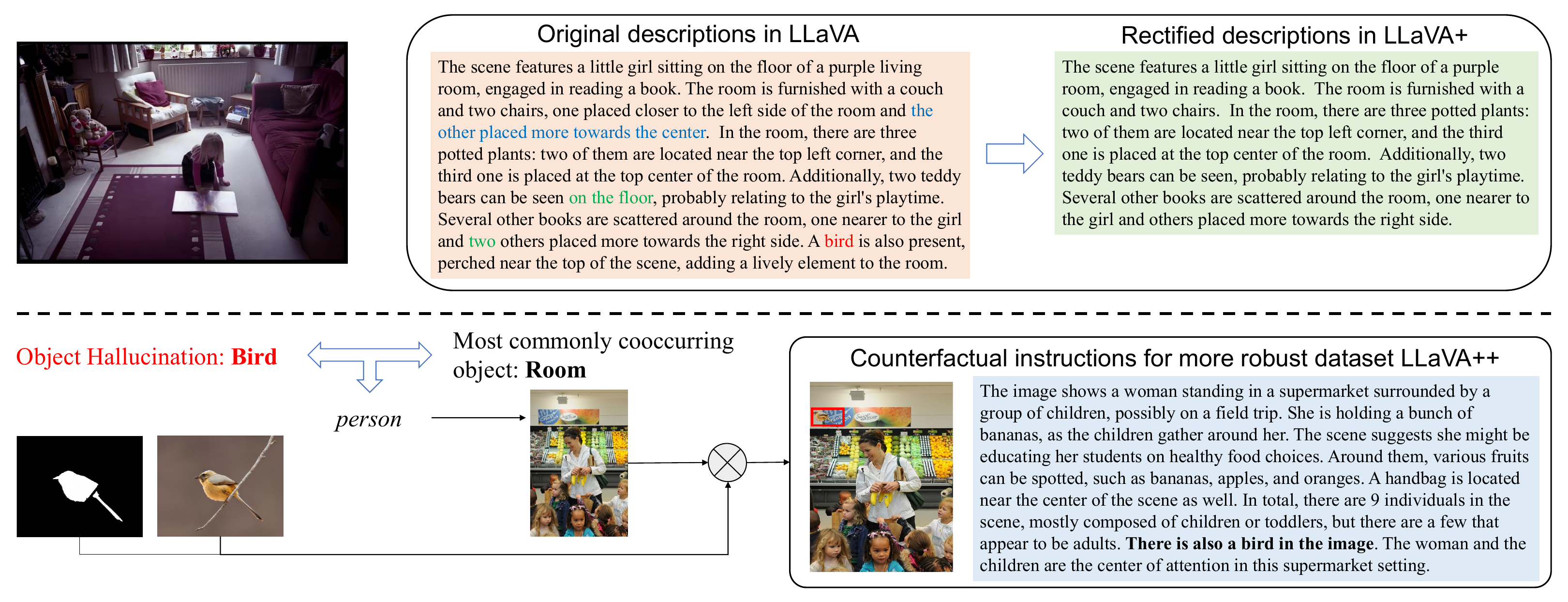}
  \caption{The instruction data visualization of LLaVA+ after hallucination elimination and LLaVA++ after visual instruction expansion.}
  \label{visualization_dataset}
\end{figure*}

\noindent\textbf{Dataset visualization.} Figure~\ref{visualization_dataset} shows some more visualized examples in the rectified dataset LLaVA+ and more robust dataset LLaVA++.

\noindent\textbf{Dataset evaluation.} Similar to Sec.~5.5, we perform a manual evaluation of the generated data for more accurate results. We sample 200 instructions from LLaVA+ and LLaVA, assessing their accuracy and quality. LLaVA+ not only shows higher accuracy scores than LLaVA~(451 v.s. 371) but also maintains comparable quality~(405 v.s. 412).
\subsection{MLLMs' Inference Analysis}
\begin{figure*}
  \centering
  \includegraphics[width=1.\linewidth]{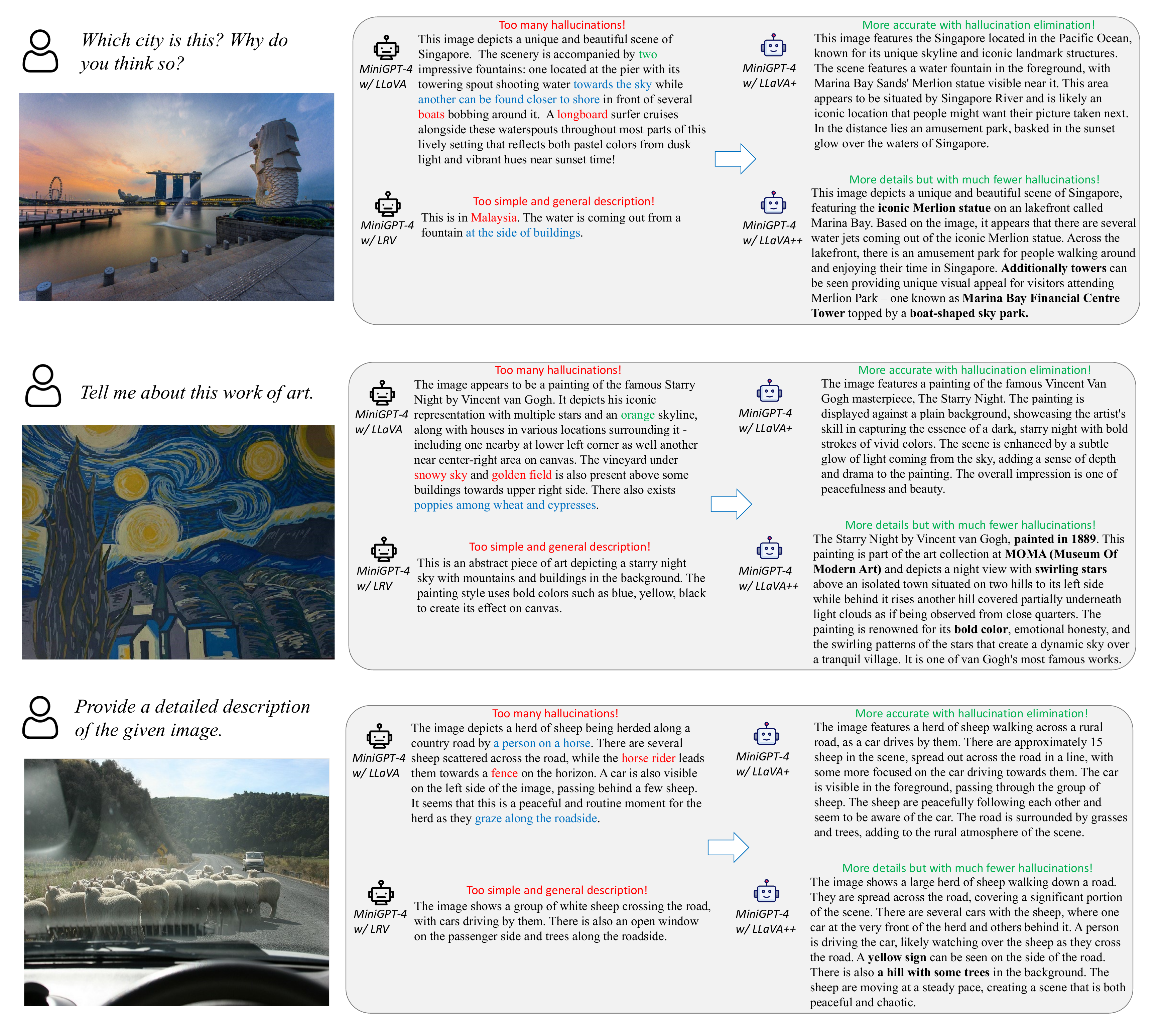}
  \caption{More visualization of MLLM comparison on various visual perception cases~(multi-round conversation, single-round conversation, detailed description, etc.).}
  \label{visualization_Result}
\end{figure*}

We compare the outputs of MiniGPT-4~\cite{zhu2023minigpt} fine-tuned on LLaVA~\cite{liu2023visual}, LRV-Instruction~\cite{liu2023aligning}, LLaVA+, and LLAVA ++ on various types of images and show the visualized results in Figure~\ref{visualization_Result}. The results verified that LLaVA+ effectively helped MLLMs eliminate hallucinatory descriptions, and LLAVA ++ further added reliable detailed descriptions.




\end{document}